%% file: main_camera_arxiv.tex
\documentclass[10pt,twocolumn,letterpaper]{article}

\usepackage{cvpr}              


\definecolor{cvprblue}{rgb}{0.21,0.49,0.74}
\usepackage[pagebackref,breaklinks,colorlinks,allcolors=cvprblue]{hyperref}

\usepackage{amsmath,amsfonts}
\usepackage{algorithm}
\usepackage{algpseudocode}
\usepackage{array}
\usepackage{textcomp}
\usepackage{stfloats}
\usepackage{url}
\usepackage{verbatim}
\usepackage{graphicx}
\usepackage{amsthm,amssymb}
\usepackage{pst-node}
\usepackage{caption}
\usepackage{bbm}
\usepackage{booktabs}
\usepackage{multirow}
\usepackage{rotating}

\newcommand{\scheme}{DeDe\xspace}

\def\paperID{11083} 
\def\confName{CVPR}
\def\confYear{2025}

\title{\scheme: Detecting Backdoor Samples for  SSL Encoders via Decoders}

\author{
Sizai Hou\textsuperscript{1}, Songze Li\textsuperscript{2,3}, Duanyi Yao\textsuperscript{1}\\
\textsuperscript{1}Hong Kong University of Science and Technology, Clear Water Bay, Hong Kong\\
\textsuperscript{2}Southeast University, Nanjing, China\\
\textsuperscript{3}Engineering Research Center of Blockchain Application, \\ Supervision and Management (Southeast University), Ministry of Education, China \\
{\tt\small shouac@connect.ust.hk, songzeli@seu.edu.cn, dyao@connect.ust.hk}
\and
}

\begin{document}
\maketitle

\begin{abstract}
Self-supervised learning (SSL) is pervasively exploited in training high-quality upstream encoders with a large amount of unlabeled data. However, it is found to be susceptible to backdoor attacks merely via polluting a small portion of training data. The victim encoders associate triggered inputs with target embeddings, e.g., mapping a triggered cat image to an airplane embedding, such that the downstream tasks inherit unintended behaviors when the trigger is activated. Emerging backdoor attacks have shown great threats across different SSL paradigms such as contrastive learning and CLIP, yet limited research is devoted to defending against such attacks, and existing defenses fall short in detecting advanced stealthy backdoors. To address the limitations, we propose a novel detection mechanism, \textbf{\scheme}, which detects the activation of backdoor mappings caused by triggered inputs on victim encoders.
Specifically, \scheme trains a decoder for any given SSL encoder using an auxiliary dataset (which can be out-of-distribution or even slightly poisoned), so that for any triggered input that misleads the encoder into the target embedding, the decoder generates an output image significantly different from the input. \scheme leverages the discrepancy between the input and the decoded output to identify potential backdoor misbehavior during inference. We empirically evaluate \scheme on both contrastive learning and CLIP models against various types of backdoor attacks. Our results demonstrate promising detection effectiveness over various advanced attacks and superior performance compared over state-of-the-art detection methods. 
\end{abstract}

\vspace{-5mm}
\section{Introduction}

Self-supervised learning (SSL), as a crucial learning paradigm, is pervasively exploited in miscellaneous domains like computer vision and natural language processing ~\cite{jaiswal2020survey, gui2024survey}. Its capability of learning without human-annotated data greatly benefits on building large foundation models and pre-training encoders via large-scale datasets. However, the obtained dataset from the Internet and the reliance on outsourced third-party models introduce enormous risks to the users, particularly in the form of backdoor attacks (or targeted attacks) \cite{chen2017targeted, gao2020backdoor, saha2020hidden}. While backdoor attacks have predominantly been studied in the context of supervised learning, there has been a notable increase in the occurrence of backdoor attacks in SSL in recent years.

In the context of a pre-trained encoder, the primary objective of an SSL backdoor attack is to match the triggered inputs toward designated embeddings. The secondary objective is to preserve the model's normal functionality.
In typical SSL paradigms, such as contrastive learning (CL) ~\cite{chen2020simple, he2020momentum} and masked autoencoders (MAE)~\cite{he2022masked}, input images are augmented in multiple views to facilitate the learning of similarities between embeddings, which significantly undermines the primary goal of backdoor attacks. But as backdoor methodologies develop, attacking CL is proved feasible by methodologies like~\cite{jia2022badencoder,zhang2024data,saha2020hidden, li2023embarrassingly}. Moreover, attacks are becoming progressively more stealthy at both the image level ~\cite{doan2021backdoor, doan2021lira} and the embedding level ~\cite{tao2024distribution}.  Image triggers are designed to be imperceptible to human inspection, while the stealthy embeddings are crafted to blend in with the target-class embeddings. This characteristic renders the detection of attacks particularly challenging.

Without any information about the pre-training dynamics and datasets, significant challenges arise for backdoor detection in an SSL encoder. Even if the poisoned pre-training dataset is given, the lack of label reference exacerbates the difficulty of detection. In the absence of any indication of backdoor existence, a model user might not realize when the hidden backdoor already affects the downstream tasks.   The challenge lies in how to detect and identify a backdoor in the SSL settings, and what is the least information a defender needs to learn. For the above reasons, there are insufficient countermeasures against backdoor attacks in CL ~\cite{feng2023decree, pan2023asset}. 
Similar to CL, CLIP is also susceptible to backdoor attacks due to its language supervision nature of maximizing the similarity between image embeddings and text embeddings ~\cite{carlini2021poisoning}. This approach allows attackers to maliciously match triggered images with target text, thereby facilitating the implantation of a backdoor~\cite{bai2024badclip, liang2024badclip}.

The scarcity and limitations of SSL backdoor countermeasures lead us to consider how a backdoor works in an SSL framework: misbehavior only happens when the trigger in input activates a victim encoder to maliciously map it to designated embedding. Consequently, we are motivated to address the challenge by monitoring potential backdoor behaviors during the inference stage. For instance, in practical scenarios where a downstream user downloads a pre-trained encoder to train their classifier, it would be beneficial to have an implemented alert system that activates if backdoor behavior is detected.  Considering that a backdoor is activated only in the presence of both a backdoor encoder and a trigger input, it raises the question: \textit{can we develop an alert system capable of detecting backdoor activation without relying on any information that might be inaccessible to an end user?}

In this paper, we propose \scheme, an effective \underline{\textbf{de}}tector for \underline{\textbf{de}}tecting backdoor samples in SSL settings. To address the limitations of existing detection methods and the emergence of increasingly advanced attacks, the proposed approach directly leverages the relationships between embeddings and images. \scheme aims to establish a region-to-region mapping to invert the normal mapping of an encoder, thereby enabling the identification of malicious behaviors that deviate from the normal mapping. 
Encoders' normal utility can be considered as the correct mapping from similar data (usually from the same class) to a certain region of the embedding space~\cite{tian2020rethinking}. Embeddings of similar images are observed to converge to a clustered region in the vector space; conversely, backdoor encoders aim to map triggered images to a different embedding region instead. 
Our approach exploits this property to
find a bijective mapping between classes of images and classes of embeddings, where normal utility leads to \textit{intra-class mapping}, while backdoor behaviors introduce
\textit{inter-class mapping}. 

Overall, the proposed detection approach offers three significant advantages: 1) the proposed approach is effective and robust against various advanced backdoor attacks, including which use the invisible trigger or stealthy embedding; 2) it does not rely on the availability of a clean dataset or an in-distribution dataset; and 3) it is a non-invasive detection method that requires no prior information regarding the victim encoder or the type of backdoor trigger, as it solely alerts in response to the occurrence of backdoor behavior. 
The contributions are summarized as follows:
\begin{itemize}
    \item We identify the pre-trained encoder mapping and its inverse's capability of detecting potential backdoor behaviors based on the intrinsic objectives of backdoor attacks.
    \item We propose \scheme as a novel backdoor detection mechanism utilizing the inverse mapping of an encoder. It trains a decoder for the inverse mapping of the encoder and reduces the reconstruction uncertainty by leveraging global embedding and constrained local image features.
    \item Comprehensive experiments are conducted to demonstrate the effectiveness of our backdoor detection method in both CL and CLIP frameworks against various attack scenarios. In comparison to existing defense approaches, our method exhibits consistent performance and achieves high detection accuracy across all circumstances. 
\end{itemize}

\vspace{-3mm}
\section{Related Works}  
\vspace{-1.5mm}
\subsection{Pre-train Encoders via SSL}
\vspace{-1.5mm}

Self-supervised learning (SSL) plays a crucial role in training large upstream encoders as it frees massive dataset label annotations from human labor and potential human errors \cite{tian2020rethinking, jaiswal2020survey, gui2024survey}. Various SSL techniques have been developed to facilitate real-world data mining, as the presence of unlabeled training instances is a common characteristic of large-scale datasets. It primarily seeks to leverage the inherent contextual relationships among the provided samples, which includes aspects such as spatial information, color information, the preservation of both local and global consistency, and other forms of prior knowledge to effectively perform a pre-training task. 
There are many different approaches to learning an encoder on an unlabeled dataset. Here, three widely recognized approaches are particularly noteworthy and relevant to our scope. Contrastive learning (CL) \cite{chen2020simple, chen2020improved, wu2018unsupervised} exploits random data augmentations to transform unlabeled images into positive pairs and optimize their similarity. Masked auto-encoder (MAE) \cite{he2022masked} is another promising approach that masks a portion of images and trains an encoder-decoder model to reconstruct the entire image. Recently, CLIP is an innovative form of CL using natural language supervision trained on a vast dataset consisting of 400 million image-text pairs sourced from the internet~\cite{radford2021learning}. Without any explicit labels, it uses images and captions, such as a sentence describing the image content, to jointly train an image encoder and a text encoder.  Its promising performance has drawn great attention into the field of multi-modal learning ~\cite{gomez2017self}. Through these  approaches, a pre-trained encoder is developed, and subsequently can be utilized for downstream supervised tasks, such as classification. 

\vspace{-2mm}
\subsection{Backdoor attacks and defences in SSL}\label{section: related works, attk and def}
\vspace{-2mm}

Since contrastive learning has been proposed, BadEncoder~\cite{jia2022badencoder} is the first to study backdoor attacks in self-supervised learning. It introduces an algorithm designed to optimize a malicious encoder by minimizing the distance between backdoor inputs and target inputs, while simultaneously preserving the effectiveness and utility of the clean encoder. The subsequent work, CorruptEncoder~\cite{zhang2024data}, investigates backdoor attacks executed solely through data poisoning. It crops the entity from the target image and overlays it onto different backgrounds to augment the image, thereby improving the effectiveness of the attack. Some attacks \cite{saha2022backdoor, li2023embarrassingly} propose approaches with additional information to inject the backdoor into the SSL model. Specifically, \cite{saha2020hidden} poisons images within a specific category and groups them during SSL training, resulting in misclassification by the downstream classifier in subsequent tasks. CTRL~\cite{li2023embarrassingly} trains the encoder with the classifier in an end-to-end manner, leading to a malicious encoder that generates shifted embeddings for the backdoor inputs.  As a counterpart to backdoor attacks, \cite{liu2022poisonedencoder} and \cite{he2022indiscriminate} investigate untargeted attacks (poison attacks) in SSL frameworks, indicating that SSL models are also vulnerable to model poisoning.

\begin{figure*}[t]
  \centering
  \begin{subfigure}{0.24\linewidth}
    \includegraphics[width=\textwidth]{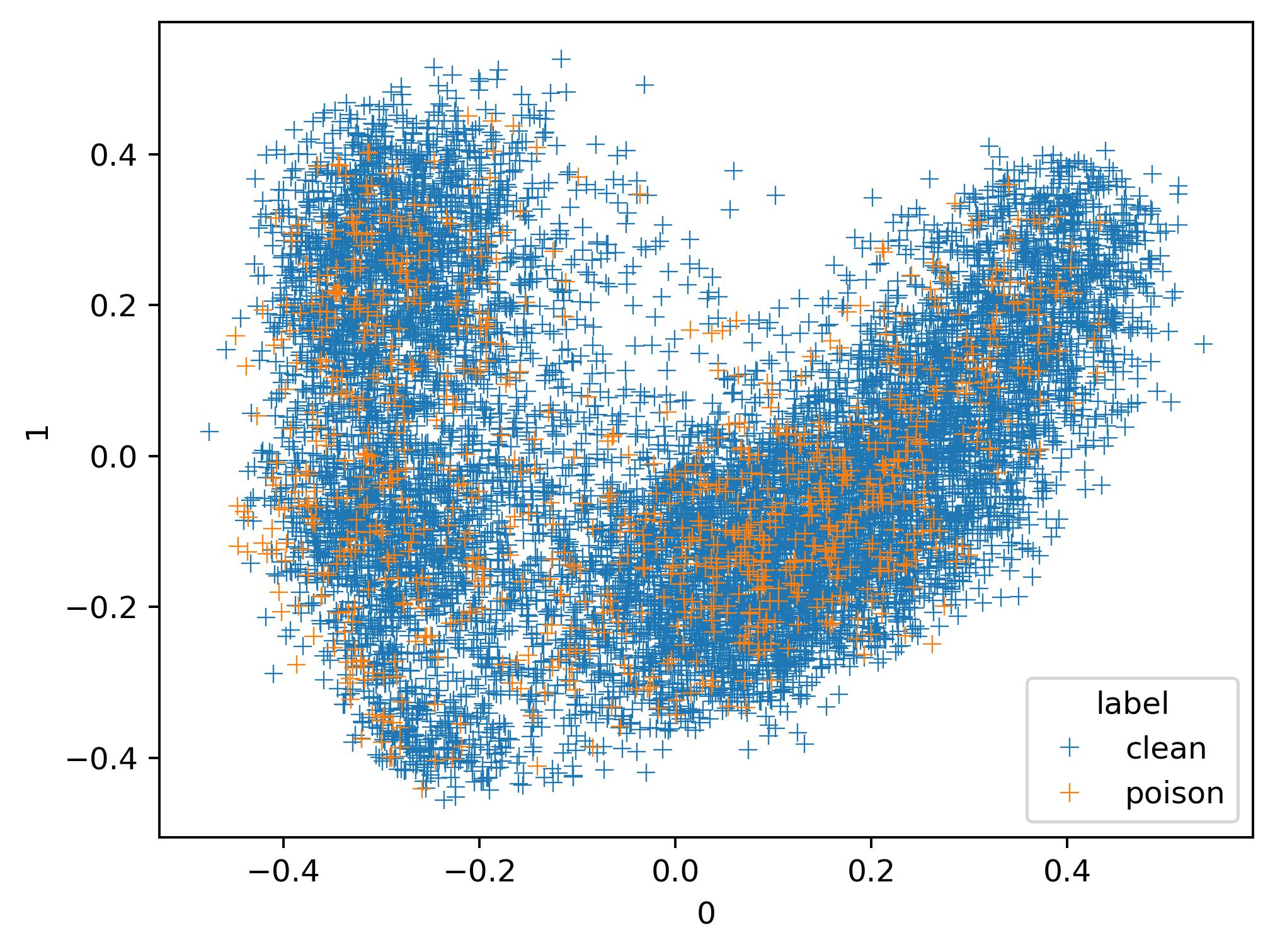}
    \caption{Clean model.}
    \label{fig: visual clean}
  \end{subfigure}
  \hfill
  \begin{subfigure}{0.24\linewidth}
    \includegraphics[width=\textwidth]{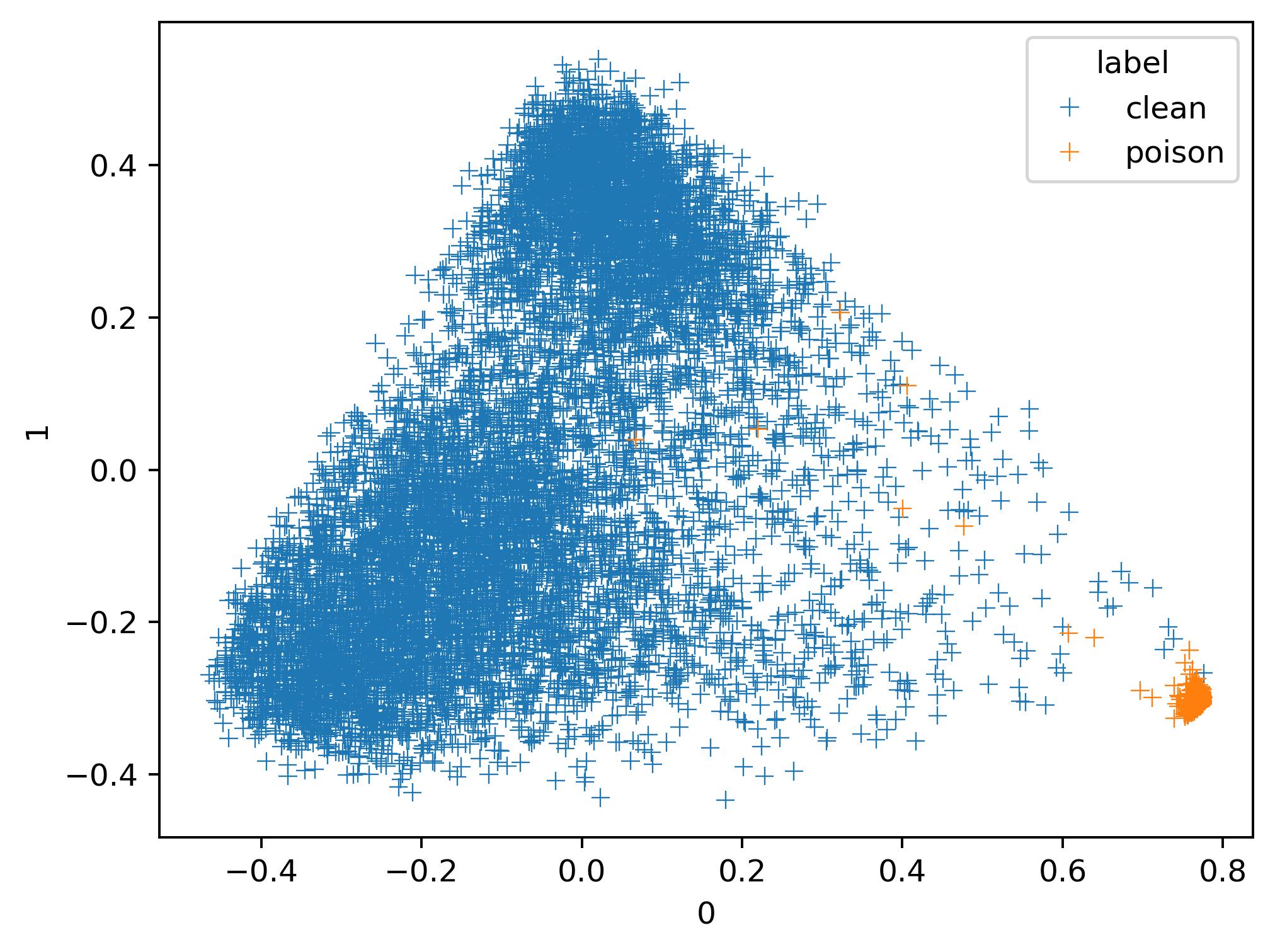}
    \caption{BadEncoder~\cite{jia2022badencoder}}
    \label{fig: visual badencoder}
  \end{subfigure}
  \hfill
  \begin{subfigure}{0.24\linewidth}
    \includegraphics[width=\textwidth]{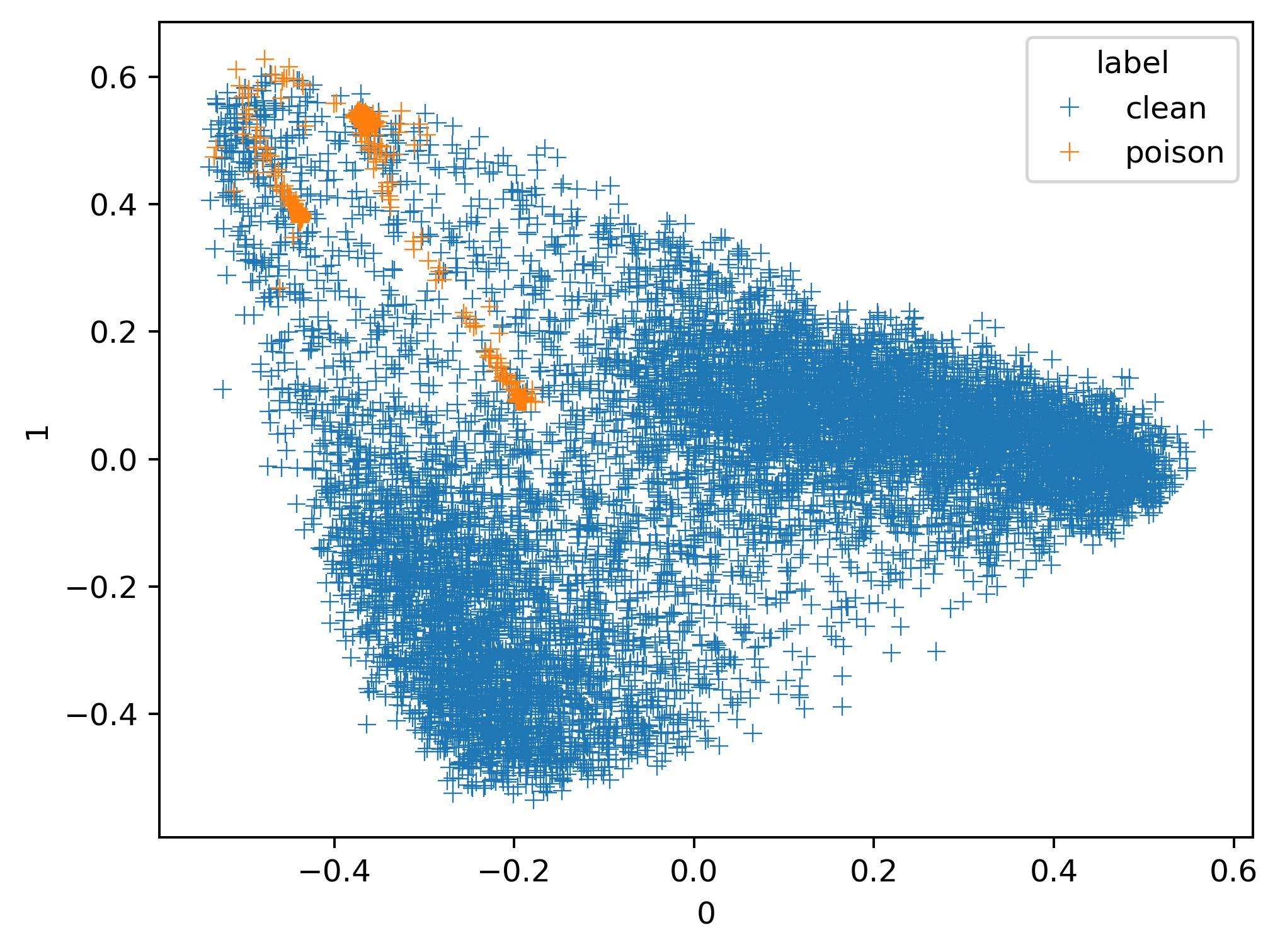}
    \caption{DRUPE~\cite{tao2024distribution}}
    \label{fig: visual drupe}
  \end{subfigure}
    \hfill
    \begin{subfigure}{0.24\linewidth}
    \includegraphics[width=\textwidth]{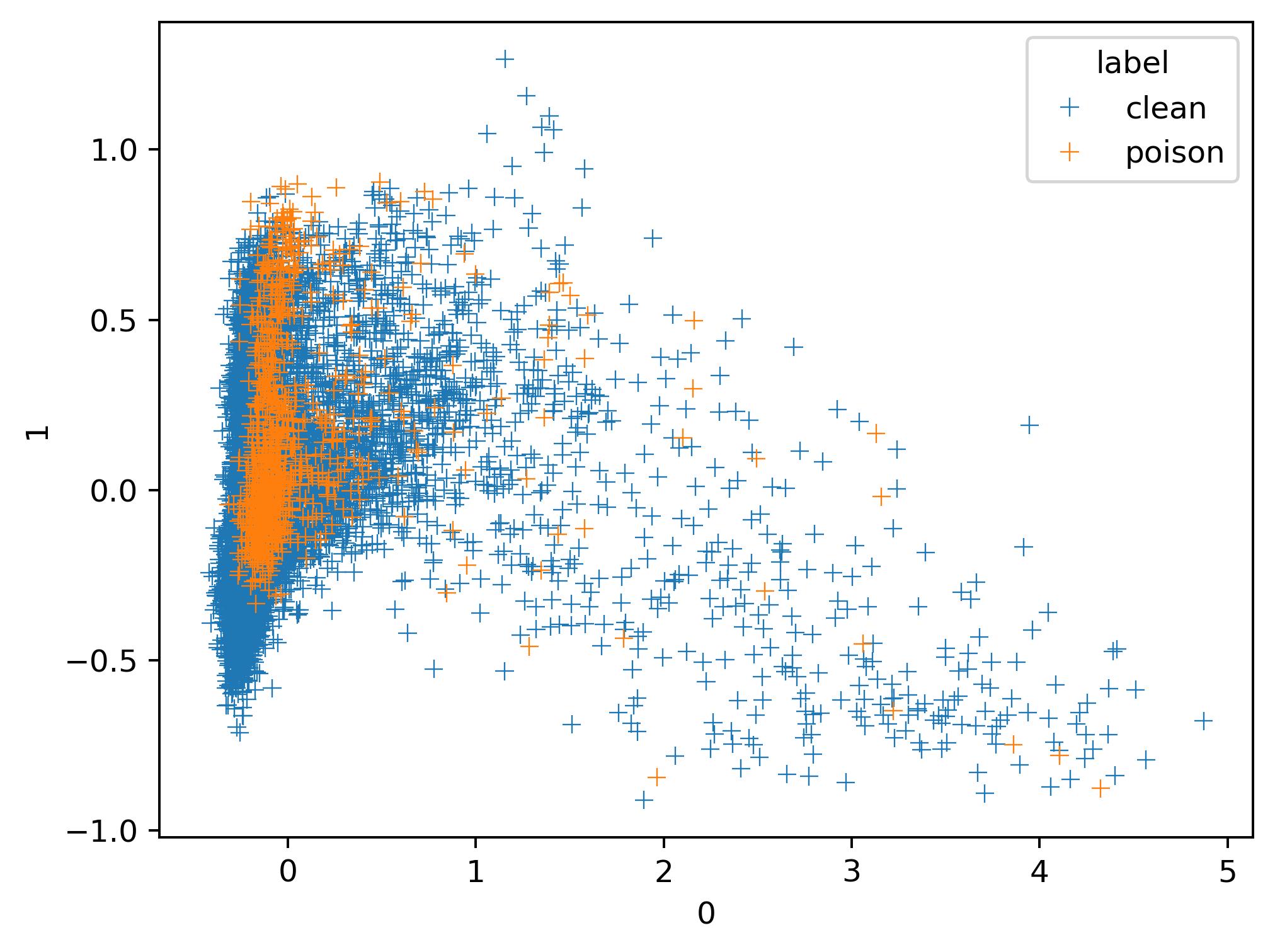}
    \caption{BadCLIP~\cite{liang2024badclip}}
    \label{fig: visual badclip}
  \end{subfigure}
  \vspace{-3mm}
\caption{Embedding Visualizations.}
\label{fig: visualization}
\vspace{-7mm}
\end{figure*}

Detecting backdoor in SSL is difficult due to the lack of data labels. There is no information available regarding which part of the data is triggered or the specific type of triggers present in the poisoned samples. This situation leads to a limited number of defensive strategies against SSL backdoor attacks. DECREE~\cite{feng2023detecting} is the first detection algorithm in the SSL settings. It requires the use of the shadow dataset, which contains backdoor samples, to optimize the trigger pattern. This is based on the observation that the distribution of clean samples and backdoor samples tends to cluster to different distributions. Subsequently, it determines whether the given encoder is implanted with a backdoor by assessing the proportion of the optimized trigger present in the entire image.
ASSET~\cite{pan2023asset} further proposes a detection method designed to identify poisoned samples in a given dataset across different learning paradigms, including SSL. It designs two opposite optimization problems over a small but clean dataset and the shadow dataset. By maximizing the optimization over the shadow dataset and minimizing over the clean dataset,  the opposite optimization offsets the effect of the clean samples, leaving poisoned samples with relatively high loss, which provides a trace for the detection. 
It is worth noting that the pipeline of backdoor sample detection like ASSET is similar to a few works in out-of-distribution (OoD) detection~\cite{hsu2020generalized, yang2022openood, yang2024generalized}, though with different objectives.
\cite{zheng2023ssl} propose to identify triggered images by detecting outliers in the embeddings via the K-means algorithm and reverse-engineers the trigger in order to mitigate the backdoor's effect. As for ~\cite{yang2023ssl}, it presents an algorithm that distills a clean encoder from a compromised encoder through Min-Max optimization with the assumption of the backdoor's existence.

CLIP~\cite{radford2021learning} is frequently regarded as a contrastive learning paradigm for multi-modal tasks, and it has also been found susceptible to the backdoor threat. It is shown the backdoor may easily be injected into the CLIP model by merely $0.0001\%$ of poisoned samples  ~\cite{carlini2021poisoning}. Sequentially, BadCLIP~\cite{liang2024badclip} proposes a novel attack that can bypass existing defenses and another BadCLIP~\cite{bai2024badclip} proposes a prompt-based backdoor attack to avoid malicious fine-tuning. 
In terms of countermeasures, there are two existing robust training frameworks; however, there are currently no detection methods effective in the presence of triggered samples within the dataset. As a robust training framework, CleanCLIP~\cite{bansal2023cleanclip} designs an additional contrastive loss for each modal. ROCLIP~\cite{yang2024robust} employs a candidate pool to match image embeddings with text embeddings at regular intervals, in which way, disrupting any potential associations between poisoned text-image pairs and realigning the correct relationships among text-image pairs. In BadCLIP~\cite{liang2024badclip}, an active implanting algorithm still succeeds in injecting the backdoor into the model. 

\vspace{-3mm}
\section{Motivations}
\vspace{-1.5mm}

\subsection{What defines a good backdoor in encoders}\label{section: what is good backdoor}
\vspace{-1.5mm}
In the early exploration of backdoor attacks in SSL, several threat models have been discussed. For example, \cite{saha2022backdoor} introduces the poisoned samples into the target category image set and CTRL~\cite{li2023embarrassingly} proposes an end-to-end backdoor attack by implanting backdoor to upstream task and downstream task together. When isolating their encoders, the backdoor samples stand out at the representation level. The threat model gets clearer when BadEncoder~\cite{jia2022badencoder} proposes the embedding-level backdoor attack where that upstream task is isolated and the triggered samples ought to mimic the target-class embedding to cause backdoor. However, in \cite{tao2024distribution}, researchers suggest that encoders generate out-of-distribution embeddings for triggered inputs, which makes it feasible to detect backdoors via defenses like DECREE ~\cite{feng2023decree}. To improve the stealthiness, \cite{tao2024distribution} presents a new attack by limiting the distributional distance between the backdoor embeddings and the clean embeddings, namely DRUPE. An increasing number of research efforts are being made towards more stealthy backdoors while effectively inducing misbehavior in models.

To testify to the assertion and substantiate the challenge of stealthy backdoors, we provide a visualization of the embeddings associated with certain SSL backdoor attacks in Figure \ref{fig: visualization} via PCA. As shown, attacks like BadEncoder and CTRL (see Figure \ref{fig: visual ctrl}) can be easily detected in the embedding space by any clustering algorithms, even for attacks imperceptible in the image space like CTRL. 
As the motivation suggested by \cite{jia2022badencoder, tao2024distribution}, an ideal backdoor in encoder misleads the downstream tasks by altering the embedding of triggered input to be close to the target-class representations. DRUPE is stealthier than its predecessors as visualized in Figure \ref{fig: visual badencoder}, \ref{fig: visual drupe} and bypasses detection methods like DECREE. Therefore, we are motivated to develop a backdoor detector that is robust to any attacking scenarios, specifically for either out-of-distribution embeddings or ideally mixed embeddings. 

\begin{figure*}[t]
  \centering
  \vspace{-6mm}
   \includegraphics[width=0.95\linewidth]{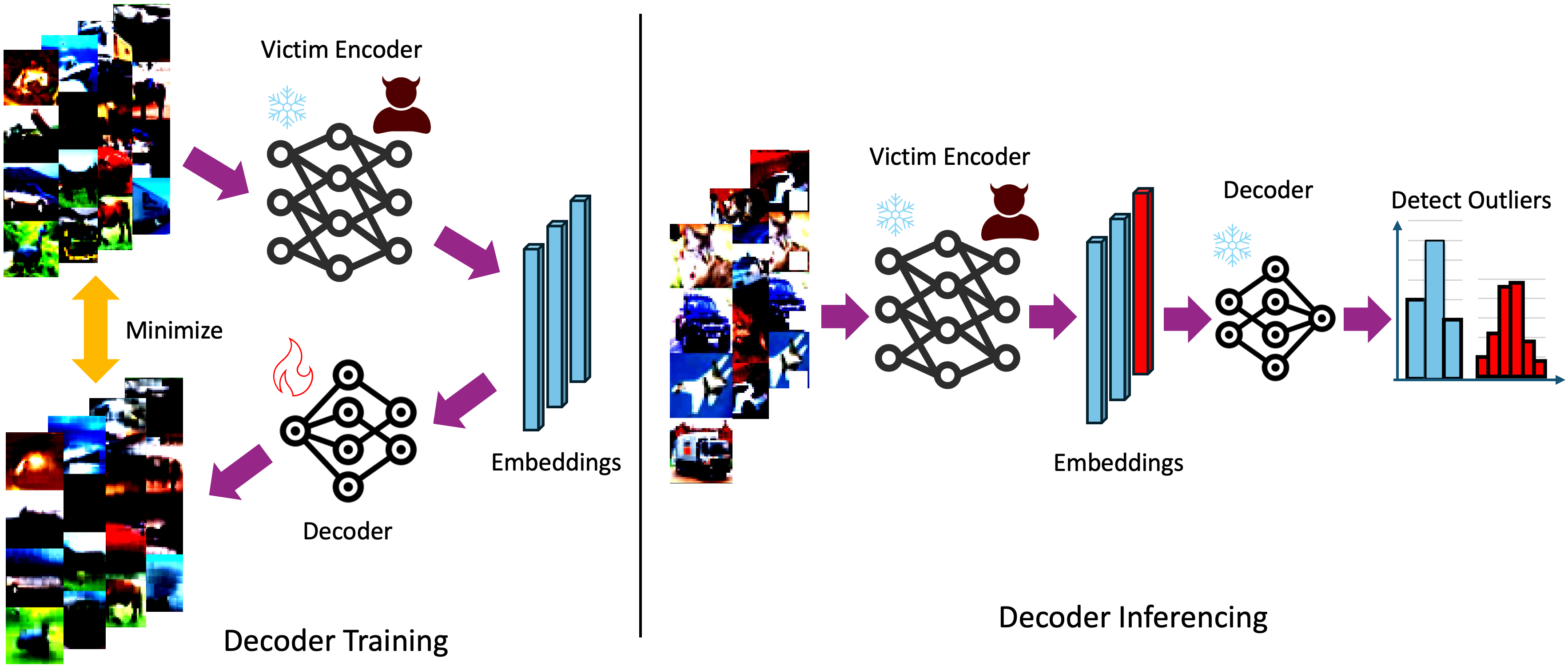}
   \vspace{-3mm}
   \caption{The Workflow of \scheme.}
   \vspace{-3mm}
   \label{fig: workflow}
   \vspace{-2mm}
\end{figure*}

\vspace{-2mm}
\subsection{Limitations in Existing Backdoor Defences}
\vspace{-1mm}
The techniques to defend against SSL backdoor attacks are still under-explored. It has been pointed out that the backdoor injected in contrastive learning is intrinsically different from its counterpart in supervised learning because the contrastive learning entangles the main task and the backdoor task, resulting in a mixture of inter-class and intra-class mapping~\cite{li2024difficulty}. DECREE as the first one to explore the defending mechanism in SSL scenario, utilizes the observation on the embedding distribution that the backdoor samples are often clustered together and highly concentrated in a certain area, therefore, finding the small perturbation that can push a clean input towards the concentration area~\cite{feng2023decree}. However, it suffers from two challenges. One is that it can only defend against attacks who demonstrate observable embedding patterns, which can be bypassed by newly developed imperceptible attacks like ~\cite{tao2024distribution}. The other challenge is that DECREE is only applicable to patch-like triggers, leading to its ineffectiveness in defending attacks such as ~\cite{li2023embarrassingly,saha2022backdoor, doan2021backdoor}. 
Another defense method worth noting is ASSET\cite{pan2023asset}, which uses the offset effect of two optimization tasks on the poisoned dataset and a manually chosen clean dataset to neutralize the clean samples influence, leaving the backdoor samples with a large loss to be detected. It has good performance in SSL attacks in general, however, it requires a sifted clean dataset from the training data, which might not always be available. It also shows a performance degeneration in more stealthy attacks and in different poison ratios. 
Moreover, there is a scarcity of effective backdoor countermeasures for SSL backdoor attacks. \cite{li2024difficulty} has shown great challenges in terms of the learning dynamics and the detection scenarios. The existing works also have their own limitations. Hence, we are motivated to develop a defense mechanism to handle various backdoor attacks with stealthy embeddings or with imperceptible triggers from a more basic perspective of the attacker's goal to manipulate the mappings of the encoder.




\vspace{-2mm}
\section{Design of \scheme}
\vspace{-1mm}

\subsection{Attack and defense models}
\vspace{-1mm}

\textbf{Attack scenario}. We consider  the two most practical scenarios from recent advancements in SSL paradigms. 

\textit{On the model provider side}, attackers may release poisoned samples to the web such that the automatically extracted data is poisoned similar to assumptions of ~\cite{saha2022backdoor, li2023embarrassingly, liu2022poisonedencoder, zhang2024data} . SSL serves as a cheap and efficient way to train encoders without human censoring, leaving a loophole for attackers to pollute the collected dataset.
Hence, access to the meta-training dataset is granted, or at least a subset of it. 

\textit{On the end user side}, consider someone who downloads pre-trained encoders to perform its own downstream tasks as an end user, which is also assumed by \cite{jia2022badencoder, liang2024badclip}. However, when the third-party service provider may have injected a backdoor into the model without affecting the main utility, how can you, as an end user, know it without any knowledge of the training dynamics or training data? Hence, no training data is known but any publicly available dataset can serve as an auxiliary dataset.

\noindent\textbf{Detection Goal}. We assume that the attacker has already implemented the backdoor into the pre-trained encoder. How the backdoor is implemented and what type of trigger is both are agnostic. As a victim encoder needs to interact with a triggered sample to cause a backdoor behavior, 
we propose a detection method, \scheme, that detects the activation of backdoor behaviors when the victim model is inferencing a triggered input to a target embedding. 

\noindent\textbf{Defender's capability}. The defender is given the victim encoder and a training dataset. The training dataset can be a subset of the encoder's pre-training dataset, which is slightly poisoned (poison rate less than $1\%$), or be a publicly available dataset without labels, such as unlabeled CIFAR10 , and STL10.

\vspace{-1.5mm}
\subsection{Key Idea}
\vspace{-1.5mm}

\begin{figure}[t]
  \centering
  \vspace{-6mm}
   \includegraphics[width=0.9\linewidth]{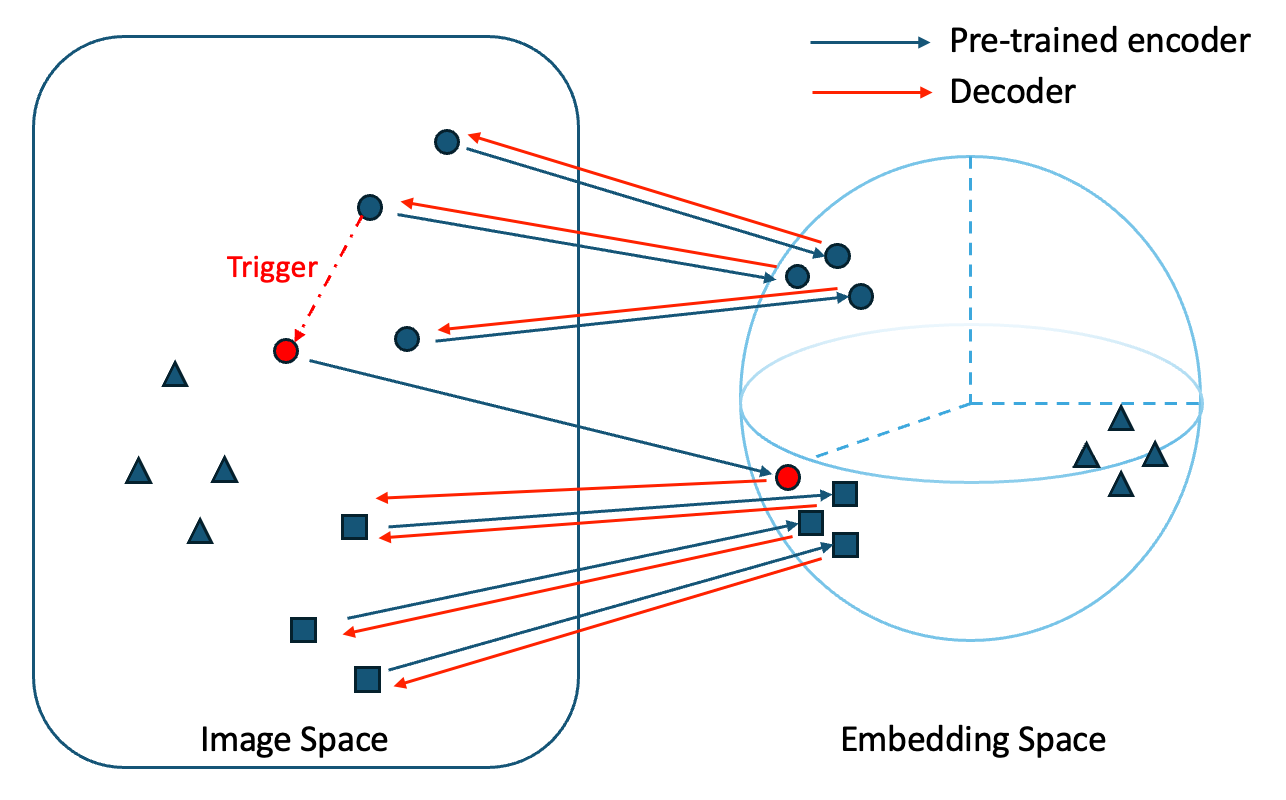}
   \vspace{-4mm}
   \caption{The Learning Goal for Decoder of \scheme.}
   \label{fig:key idea}
   \vspace{-7mm}
\end{figure}

In essence, SSL spontaneously maps images with similar features to clusters, which is determined by its training process. We present t-SNE plots for different SSL paradigms in the Appendix.
However, backdoor mappings force cross-class mapping from a class of image to the target-class embedding, which is illustrated by the red dots in Figure \ref{fig:key idea}. As required by backdoor's objective, an ideal backdoor should remain imperceptible at both the image level and the embedding level. At the image level, the attacker applies only a small perturbation (trigger) as a special distribution in image space, avoiding possible human inspections. At the embedding level, the attacker forces the victim encoder to learn the particular mapping for triggered samples to the targeted embedding while keeping clean images to the respective embeddings as the blue dots in Figure \ref{fig:key idea}. 
A naive instinct is that embeddings from the same cluster should represent similar visual features in images. 
If we learn the reversed mapping from embedding clusters to images as the red arrows in the figure, the reversed mapping should project certain embedding regions to certain visual features in the image, then the backdoor mapping is unlikely to be preserved. It is because, first, the image space is much more sparse than the embedding space, so condensed embedding is unlikely to preserve a backdoor distribution in image. Secondly, since the embedding distribution forms clusters, similar embeddings tend to map to similar image features, but the backdoor feature is usually imperceptible and dissimilar to image features. 
For example, a cat image with the trigger is maliciously mapped to an embedding in the airplane's region, if we find a point around its embedding and look for the point's pre-image, it is highly probable that the corresponding image will depict an airplane.

The idea of reversing the mapping is simple, however, finding such an inverse is challenging,  because the inverted mapping is from a low-dimensional space to a high-dimensional space. The reconstruction is highly unpredictable since much information has degenerated by the encoder's dimension reduction. Thus, how to reconstruct an image from the condensed embedding information is crucial. 
Inspired by MAE~\cite{he2022masked}, we can allow some local information in image space to aid the reconstruction, which can give valuable reference for the object's color, texture, background and other local details. 
More critically, a triggered image is almost identical to the original image as the perturbation has to be small, as a result, the trigger is usually incomplete or neglected when we constrain limited information from the image space by randomly selecting small patches. 
Therefore, we divide the reconstruction dependency into two parts, one is the global embedding and the other is the local patches. Leveraging the information from both global and local embeddings, we can train a decoder that reconstructs the image to approach the original one. Calling back to the above example, we reconstruct the pre-image from the embedding around the airplane region together with a few local patches of the original cat image. What would the reconstructed image be like? The decoder might reconstruct a cat-colored airplane or get confused to produce nonsensical prediction.
In either way, when we compare the reconstruction with the actual pre-image, they are inconsistent and distinguishable by simply evaluating an error. We present the workflow of \scheme in Figure \ref{fig: workflow}.

\vspace{-2mm}
\subsection{Methodology}
\vspace{-2mm}

Given an encoder architecture $f$ and pre-trained parameters $\theta$, it is pre-trained to map any data points $\mathbf{x} $ from image space $ \mathcal{X}$ to the embedding space $\mathcal{I}$, $\mathbf{e}  = f_{\theta} (\mathbf{x} )$, facilitating any downstream tasks. Assuming the downstream task is classification, we denote a simple classifier (linear probe) by $g(\cdot)$ that maps the embedding to the class distribution $\hat{y}$. A malicious attacker manipulates parameters $\theta^*$ such that $f_{\theta^*}$ maps a triggered data point to the poisoned embedding $\mathbf{e} ^* = f_{\theta^*} (\mathbf{x} ^*)$, where $ \mathbf{x} ^* = \mathbf{x}  \oplus \mathbf{t}$ denotes the trigger pattern $\mathbf{t}$ is applied on the image. Notice that a backdoor is only activated when both input and model are backdoored, i.e. $\theta^*$ and $\mathbf{x} ^*$ present together. 
To summarize the victim encoder's behavior, we have the encoder's normal utility as $f_{\theta^*}(\mathbf{x} ) = f_{\theta}(\mathbf{x} ) = \mathbf{e} $ and backdoor behavior as $f_{\theta^*}(\mathbf{x} ^*) = \mathbf{e} ^*$ while $f_{\theta}(\mathbf{x} ^*) = \mathbf{e} $.
Denote the training dataset by $\mathcal{D}$, which can be a subset of the meta-training dataset or an independent clean dataset by its availability.

\scheme aims to train a decoder with an aiding patch encoder, denoted by $h_d,h_e$ respectively. Given pre-trained encoder $f:\mathcal{X} \mapsto \mathcal{I}$, the proposed decoder tries to inverse the mapping: $h_d:\mathcal{I} \mapsto \mathcal{X}$. For $\mathbf{x} \in \mathcal{X}$, \scheme generates a patched mask $M_\alpha$ according to the masking ratio $\alpha$, and then masks the input image $M_\alpha\odot  \mathbf{x} $, where $\odot$ denotes pixel-wise product. The higher $\alpha$ is, the more information of $\mathbf{x} $ is hidden, therefore, if $\alpha=1$, $M_{1.0}\odot  \mathbf{x}  = \varnothing$. Patch encoder $h_e(M_\alpha\odot\mathbf{x} )$ encodes the masked input to patch embeddings $\mathbf{b} $ with positional mask information. The decoder $h_d(\mathbf{e} , \mathbf{b} )$ decodes the global embedding and the patch embedding to an image reconstruction $\hat{\mathbf{x}} $. The optimization problem is formulated by minimizing:
\vspace{-2mm}
\begin{equation} \label{eq1}
\vspace{-2mm}
\begin{split}
     \mathcal{L}_{}(h_d, h_e) &= \mathop{\mathbb{E}}_{\mathbf{x}\in \mathcal{D}} || \mathbf{x}- \hat{\mathbf{x}} ||_2^2, \\
    {where}\quad \hat{\mathbf{x}} & = h_d\left ( f_{\theta^*}(\mathbf{x}), h_e(M_\alpha\odot\mathbf{x}) \right ) .
\end{split}
\end{equation}

The testing stage of \scheme is calculating the loss the same as \eqref{eq1} on the test dataset. The only difference is that the choice of $\alpha$ in the testing stage is higher than in the training stage, in order to constrain the image space information. For example, $\alpha = 0.9$ is used when training, then $\alpha = 0.99$ is used in testing. It is to force the reconstruction more depending on the global embedding to detect malformed ones.  
The testing metric is simple as follows: 
\vspace{-2mm}
\begin{equation}
    \ell\left ( \mathbf{x} \right ) = \left \| \mathbf{x} - h_d\left ( f_{\theta^*}(\mathbf{x}), h_e(M_{\min \left ( 1.1*\alpha, 1.0 \right ) }\odot\mathbf{x}) \right ) \right \| _2^2.
\end{equation}
The reconstruction error $\ell\left ( \mathbf{x} \right )$ is compared with a threshold $\tau$ to decide whether the inference is backdoor behavior. To tolerate normal reconstruction error, the threshold is set to be $1.5 \times $ average training loss. 
The training algorithm is summarized as Algorithm \ref{alg: main}, where $\theta_h$ represents the trainable parameters in $h_d, h_e$.

\begin{algorithm}[tb]

\caption{\scheme: Training the Decoder}\label{alg: main}
\hspace*{\algorithmicindent} \textbf{Input}: $\mathcal{D}, f_{\theta^*}$, any sample $\mathbf{x}_t$ from the test dataset \\
\hspace*{\algorithmicindent} \textbf{Output}: $h_e, h_d$, prediction $\hat{y}_t$ \\
\hspace*{\algorithmicindent}\textbf{Parameters}: the number of iterations $N$, learning rate $\eta$, masking ratio $\alpha$.
\begin{algorithmic}

\For{\textit{each iteration $j$}$\in 1,\ldots,N$}
        \State $\mathbf{e}_i \gets f_{\theta^*} (\mathbf{x}_i),\quad \forall \mathbf{x}_i \in\mathcal{D}$  
        \State $\mathbf{b}_i \gets h_e(M_\alpha\odot\mathbf{x}_i ),\quad \forall \mathbf{x}_i \in\mathcal{D}$ 
        \State $\hat{\mathbf{x}}_i \gets h_d(\mathbf{e}_i , \mathbf{b}_i ),\quad \forall \mathbf{x}_i \in\mathcal{D}$
        \State $\mathcal{L} \gets \frac{1}{|\mathcal{D}|}\sum_{\mathbf{x}_i \in\mathcal{D}} \left \| \hat{\mathbf{x}}_i -{\mathbf{x}}_i  \right \|^2_2$
        \State $\theta_h^{(j+1)} \gets \theta_h^{(j)} - \eta\cdot\frac{\partial\mathcal{L}}{\partial \theta_h^{(j)} } $
  \EndFor
  \State $\tau \gets 1.5\cdot\frac{1}{|\mathcal{D}|}\sum_{\mathbf{x}_i \in\mathcal{D}} \left \| \hat{\mathbf{x}}_i -{\mathbf{x}}_i  \right \|^2_2 $
  \State $\alpha \gets \min \left ( 1.1\cdot\alpha, 1.0 \right )$
  \State $\hat{y}_t \gets \mathbbm{1} ( \left \| \mathbf{x}_t - h_d\left ( f_{\theta^*}(\mathbf{x}_t), h_e(M_\alpha\odot\mathbf{x}_t) \right ) \right \| _2^2 - \tau) $

\end{algorithmic}


\end{algorithm}

\vspace{-2.5mm}
\section{Evaluation} 
\vspace{-2mm}

We evaluate \scheme in representative SSL paradigms: contrastive learning and CLIP. Their tasks are different as contrastive learning uses unlabeled image datasets for training encoder while CLIP uses image-text pair datasets to train a vision encoder and a text encoder. We use SimCLR~\cite{chen2020simple} to perform contrastive learning and use common training for CLIP~\cite{radford2021learning}. As we only consider attacks in the vision encoder, the used text encoder is publicly available on OpenAI 
\footnote{https://github.com/openai/CLIP/}. Our code is publicly available at this repository \url{https://github.com/tardisblue9/DeDe}.

We investigate various attacks and compare the proposed detection with SOTA detections to show how effective the detection \scheme is. We test these attacks including CL and CLIP attacks as well as different trigger types in different settings to show how robust the detection method is.

\vspace{-2.5mm}
\subsection{Experiment Set-up}
\vspace{-2mm}

\begin{table*}[t]
\vspace{-8mm}
\caption{Upstream Detection Performance (* is used to denote a successful detection for DECREE.)}
\vspace{-3mm}
\label{table: upstream}
\centering
\scriptsize

\begin{tabular}{cccccccccccccccc}\toprule
& \multicolumn{3}{c}{BadEncoder} & \multicolumn{3}{c}{CTRL}& \multicolumn{3}{c}{DRUPE}& \multicolumn{3}{c}{CLIP Backdoor}& \multicolumn{3}{c}{BadCLIP}
\\\cmidrule(lr){2-4}\cmidrule(lr){5-7}\cmidrule(lr){8-10} \cmidrule(lr){11-13} \cmidrule(lr){14-16}
           & \textit{TPR} $(\uparrow)$  & \textit{FPR}$(\downarrow)$ & \textit{AUC}$(\uparrow)$    & \textit{TPR}  & \textit{FPR} & \textit{AUC} & \textit{TPR}  & \textit{FPR} & \textit{AUC}& \textit{TPR}  & \textit{FPR} & \textit{AUC} & \textit{TPR}  & \textit{FPR} & \textit{AUC} \\\midrule
DECREE    & 46.4$^*$ & 70.6$^*$ & 0.362$^*$ & - & - & -  &  44.7 & 70.6 &  0.344  & 30.1$^*$ &  54.8$^*$ &  0.348$^*$  & - &  -  & - \\
ASSET & 84.9  & 4.0  & 0.978 &  89.6 &  30.2 &  0.799 &  94.7  & 27.6 &  0.858 &  30.0 &  0.0  & 0.555  & 99.8  & 49.4  & 0.773 \\
DEDE & 93.1 &  6.9  & 0.981 &  87.2 &  12.8  & \textbf{0.912} &  97.6  & 2.4 &  \textbf{0.997} &  100.0 &  0.0  & \textbf{1.0 }&  85.0  & 14.9 &  \textbf{0.925}\\
DEDE OoD & 97.2 &  2.83  & \textbf{0.993}  & 88.1  & 13.3 &  0.911  & 92.1  & 7.9  & 0.976 &  100.0 &  0.0  & \textbf{1.0}  & 72.1 &27.9 & 0.798 \\\bottomrule
           
\vspace{-5mm}
\end{tabular}

\end{table*}




\textbf{Attack models}. We choose several representative and SOTA attacks in SSL to acquire the victim encoder. They are BadEncoder~\cite{jia2022badencoder}, CTRL~\cite{li2023embarrassingly}, DRUPE~\cite{tao2024distribution}, CLIP-Backdoor~\cite{carlini2021poisoning} and BadCLIP ~\cite{liang2024badclip}. The first three are attacks on CL and the last two are attacks on CLIP. BadEncoder, DRUPE, and CLIP-Backdoor use patch-like triggers while CTRL and BadCLIP use invisible(imperceptible) triggers, specifically frequency-based and adaptive triggers respectively.   We use their public repositories to recover the attacks and evaluate their functionality on a uniform downstream baseline to ensure the clean performance and the backdoor performance, except for CLIP Backdoor for unavailability, which we use the implementation from \cite{feng2023decree}. 

\noindent\textbf{Detection models}. There are two SOTA detection methods in SSL to our knowledge: DECREE\cite{feng2023detecting} and ASSET\cite{pan2023asset}. For settings, ASSET's training requires a small but clean dataset, referred to as the base dataset,  while their test criteria are identical to ours, which is convenient to compare with. DECREE optimizes the trigger pattern with a shadow dataset, subsequently employing a criterion on the optimized trigger pattern to identify backdoor behaviors. However, it fails to detect any backdoor samples during testing.  Therefore, we compare the optimized trigger and the poisoned dataset, and calculate their similarity to detect any backdoor samples. Specifically,  we assess the $\ell_2$-norm between the optimized trigger and the image data located within the trigger region, and use the norm as the error measurement. Utilizing this error for each data point enables us to compute all scores accordingly. For DECREE, we use $*$ on scores for successful detection of the model backdoor. 

\noindent\textbf{Datasets}. As different stages in the pipeline are involved, we divide the dataset into the following categories: {pre-train dataset, shadow dataset, memory dataset, \scheme training dataset, downstream training/testing dataset}. Unified augmentation is used to rectify different color distributions. 

Since we mainly use pre-trained model checkpoints for our implementation, \underline{pre-train dataset} is omitted. For \underline{shadow dataset}, we follow attacks' public implementations, so we simply mention their datasets usage here. CIFAR-10~\cite{krizhevsky2009learningcifar10} is modified as the backdoor training dataset for BadEncoder, DRUPE and CTRL and ImageNet~\cite{deng2009imagenet} is used for backdoor injection in CLIP Backdoor and BadCLIP, whose clean models are trained on the CC3M dataset~\cite{sharma2018conceptual}. For \underline{memory dataset}, it is the clean version of shadow dataset. In ASSET, it uses a subset of memory dataset. For \underline{\scheme training dataset}, we either use a subset from the same dataset in backdoor training, i.e. CIFAR-10 for CL and ImageNet for CLIPs, or an out-of-distribution dataset (OoD) STL-10 \cite{coates2011analysisstl10} consistently. As we mentioned in Section \ref{section: what is good backdoor}, whether the \underline{downstream training data} is poisoned is crucial to some attacks. We consider datasets CIFAR-10, GTSRB~\cite{Houben-IJCNN-2013} and SVHN~\cite{netzer2011reading}, and two cases of each dataset by clean and poisoning with $1\%$ backdoor samples. The \underline{downstream testing datasets} are simply the clean test dataset and backdoor test dataset for evaluation. 

\begin{table*}[t]
\caption{Downstream Performance}
\vspace{-3mm}
\label{table: downstream}
\centering
\scriptsize
\begin{tabular}{ccccccccccccc}\toprule
& \multicolumn{2}{c}{No Attack} &  \multicolumn{2}{c}{BadEncoder} & \multicolumn{2}{c}{CTRL}& \multicolumn{2}{c}{DRUPE}& \multicolumn{2}{c}{CLIP Backdoor}& \multicolumn{2}{c}{BadCLIP}
\\  \cmidrule(lr){2-3} \cmidrule(lr){4-5} \cmidrule(lr){6-7} \cmidrule(lr){8-9} \cmidrule(lr){10-11} \cmidrule(lr){12-13} & \textit{CA} $(\uparrow)$ & \textit{ASR} $(\downarrow)$ & \textit{CA} & \textit{ASR}& \textit{CA} & \textit{ASR}& \textit{CA} & \textit{ASR}& \textit{CA} & \textit{ASR}& \textit{CA} & \textit{ASR}\\\midrule
 No Def. Poison dataset &  90.5 & 18.8 & 86.3 & 100.0 & 73.9&  95.4 & 85.6&  100.0 & 79.4 & 100.0 & 82.8 & 98.3 \\
No Def. Clean dataset & 90.7  & 10.2 &  86.4  & 99.9 &  74.0 &  63.8 &  85.8 &  96.3 &  79.7  & 98.4  & 83.1  & 11.7 \\
ASSET & -  & - &  86.3  & 55.3 &  72.7 &  49.3  & 85.9  & 53.7  & 78.5  & 98.3  & 81.1 & 97.4 \\
 DEDE & -  & -  & 86.5 &  \textbf{1.3}  & 72.2 &  \textbf{4.8}  & 85.9  & \textbf{2.1}  & 79.7  & \textbf{0.0} & 81.6 & \textbf{30.5}
    \\\bottomrule
\end{tabular}

\vspace{-5mm}
\end{table*}

\noindent\textbf{Metrics}. To compare the detection performance, we mainly use three related metrics: true positive rate (TPR), false positive rate (FPR) and area under the ROC curve (AUC score). The detection of backdoor samples is treated as a binary classification problem. Typically, the poisoned samples are in the minority to evade detection and scrutiny. TPR,  also referred to as Recall, is calculated by $TP/(TP+FN)$, which measures the accuracy of correctly detected backdoor samples. FPR compensates the TPR by reflecting the clean samples that are misclassified as backdoor, formulated by $FP/(FP+TN)$. For imbalanced binary data distribution, it is common to measure TPR, FPR as used in \cite{feng2023detecting, pan2023asset}.  But the (TPR,FPR) pair only reflects a single point on the ROC curve with a fixed threshold. To authentically represent the detection result independently from the threshold, the ROC curve is plotted by sliding the threshold. The AUC score measures the area under ROC curve and distributes in $(0,1]$, where $1$ stands for a perfect classifier. Since TPR and FPR represent a trade-off score depending on a threshold, we present the default result by the scikit-learn library~\cite{scikit-learn}.

Following the prior work \cite{pan2023asset}, we evaluate the detection performance also by the downstream tasks after removing the detected points. The metrics are the clean accuracy (CA) for the main task and the attack success rate (ASR) for the backdoor task. CA reflects the model utility of embedding, which means the higher the CA is, the better the encoder performs. ASR reflects the backdoor effectiveness, which means the higher the ASR is, the more successful the backdoor is. As explained in the dataset part, the downstream setting in attacks is relevant. So for detection settings, we set all downstream training dataset to be poisoned by $1\%$ as the challenge for detectors. Detection methods are used to filter the training and testing dataset before actual training and testing. The detected testing data are counted as negative samples in evaluating ASR. In this case, high CA indicates enough clean samples are kept for the main task training and low ASR indicates backdoor samples are successfully filtered by the algorithm.

\noindent\textbf{Model architectures}. As \scheme has no dependency on the victim encoder, we use default backbone architectures for encoders, namely ResNet-18 for BadEncoder, DRUPE, CTRL, and ViT-B/16 for CLIPs. For \scheme, we use the lightweight MAE decoder developed from ViT-B/16~\cite{alexey2020image} as described in ~\cite{he2022masked}. Our default setting only has 6 blocks for the encoder and 4 blocks for the decoder, which has a width of 512 for ResNet and 1024 for ViT. The default training mask ratio $\alpha$ is set to be $0.9$ with an exception for CTRL. The default patch size is chosen between 4 to 32 for image size from $32\times 32$ in CL and $224\times 224$ in CLIP. For datasets inconsistent with the image input size (such as STL-10), up/down sampling are used respectively.

\begin{figure}[tb]
    \centering
    \vspace{-5mm}
  \begin{subfigure}{0.48\linewidth}
    \includegraphics[width=\textwidth]{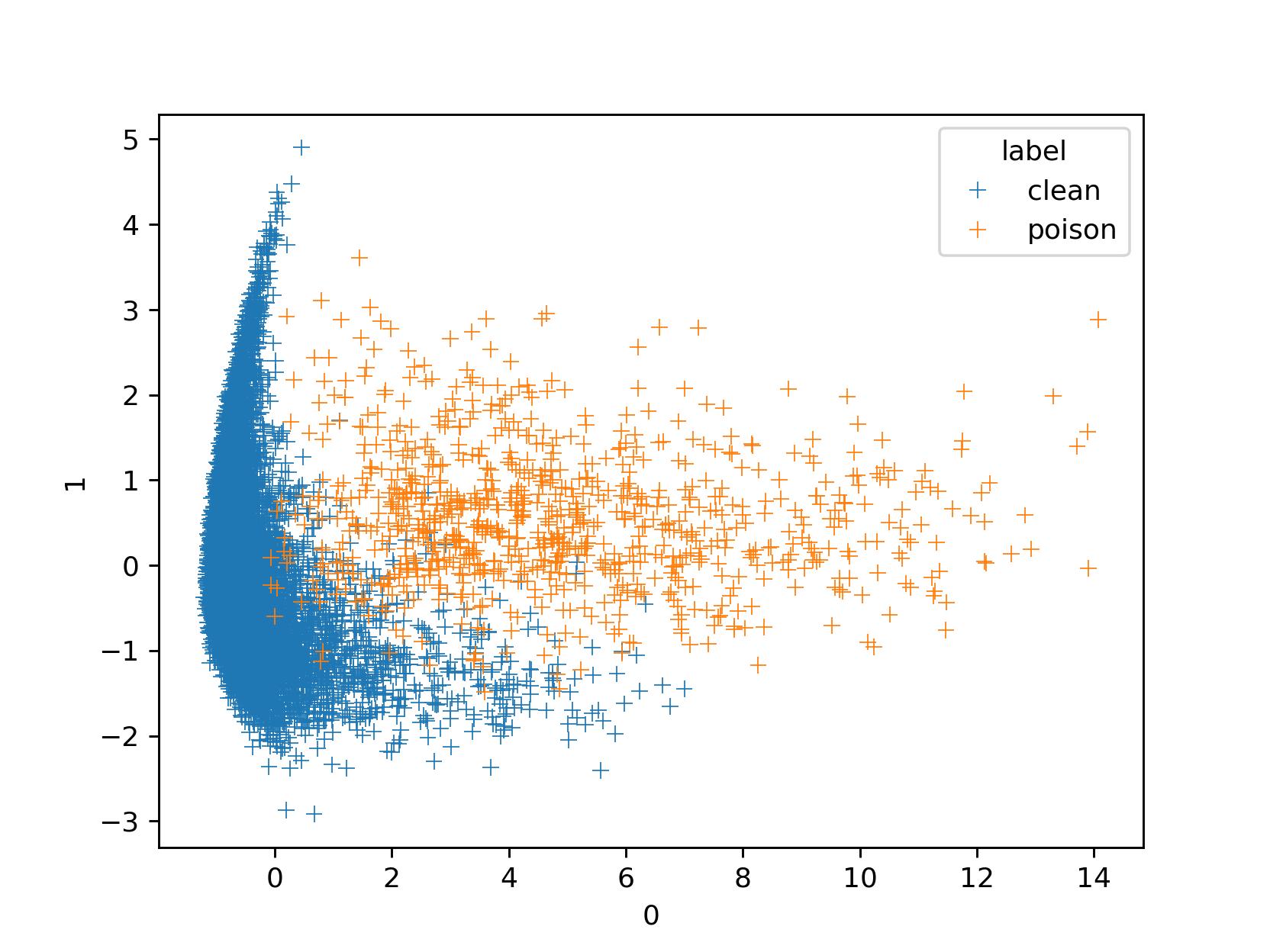}
    \vspace{-5mm}
    \caption{CTRL embedding.}
    \label{fig: visual ctrl}
  \end{subfigure}
  \hfill
  \begin{subfigure}{0.48\linewidth}
    \includegraphics[width=\textwidth]{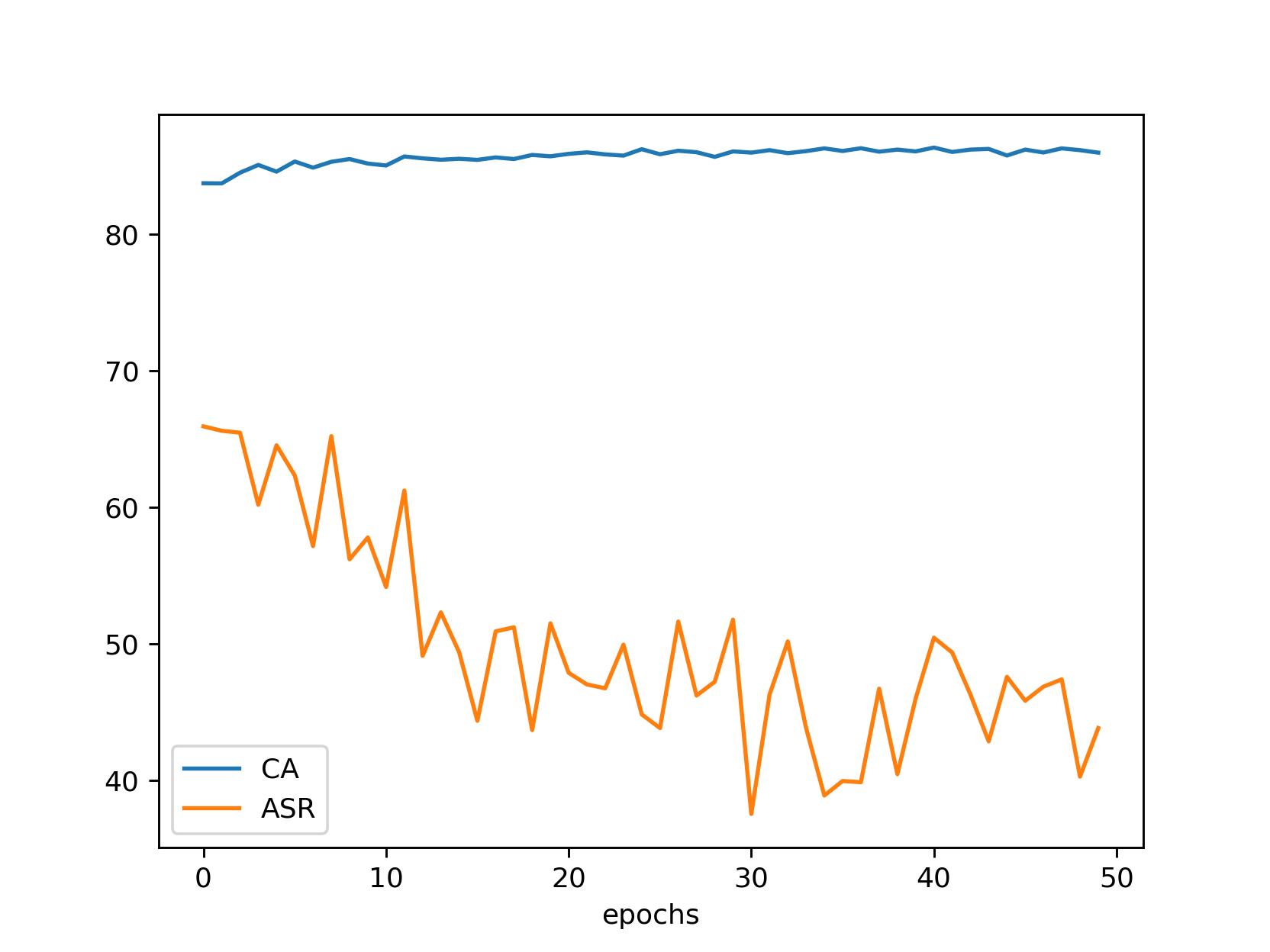}
    \vspace{-5mm}
    \caption{CTRL downstream task.}
    \label{fig: ctrl train}
  \end{subfigure}
   \label{fig: ctrl}
   \vspace{-2mm}
   \caption{Discussion for CTRL Attack}
   \vspace{-6mm}
\end{figure}

\noindent\textbf{Auxiliary Explanation.} Recall in Section \ref{section: what is good backdoor}, we define a good backdoor in SSL to be imitative mapping behavior towards the target embeddings, which is independent of the downstream classifier. 
CTRL falls short in such a definition for its end-to-end backdoor training. Hence, it embeds the triggered inputs in the out-of-distribution region across the embedding space as illustrated in Figure \ref{fig: visual ctrl}. One direct consequence is that, if we use a clean training dataset in the downstream task, its ASR drops in the training iterations as in Figure \ref{fig: ctrl train}.  Despite this, we report the highest ASR for CTRL in Table \ref{table: downstream}. When training \scheme against CTRL, we address the issue by filling the out-of-distribution regions with random noise image embeddings, where we define the region outside a norm ball with radius of 0.99 quantiles in embedding norms. Furthermore, the training masking ratio for CTRL is set at $0.75$ to provide additional support from the image space, as the embeddings are significantly sparse.

\vspace{-1mm}
\subsection{Experiment Results}
\vspace{-1mm}

We assess detection performance in two scenarios: one with a balanced data distribution in which 50\% of the data is triggered, and another with an unbalanced data distribution where only 1\% of the data is triggered (see Appendix). For ASSET, a clean subset is sampled for their model training, whose size is $5$k for CL and $500$ for CLIP. We observe that ASSET's performance is unstable when using nested offsets (epoch number is set to $10$), 
the best epoch is reported. 

In Table \ref{table: upstream}, we present the detection results of balanced data distribution. DECREE is only applicable to attacks with patch-like triggers, so it fails for CTRL with spectral trigger and BadCLIP with random trigger. Generally, \scheme has a consistent performance across different attacks with high AUC scores. In comparison to other attacks, \scheme demonstrates nearly perfect detection of the BadEncoder and DRUPE attacks, while experiencing a decline in performance with the CTRL and BadCLIP attacks, but it still upholds over 0.9 AUC scores.  ASSET also exhibits degeneracy in the cases of CTRL and BadCLIP and has a high AUC for BadEncoder.  Its performance experiences a decline due to the stealthiness of DRUPE and fails in CLIP Backdoor. \scheme maintains a high level of performance with DRUPE in comparison to BadEncoder, showing that  it is robust to the stealthiness of the embedding.

In Table \ref{table: downstream}, as we discussed in Section \ref{section: what is good backdoor}, even the clean baseline shows an increase in ASR when the downstream task is trained on a contaminated dataset. Consistently, the no-defense baselines (No Def.) always have higher attack success rates when the poisoned training dataset is used instead of a clean one.  This trend is particularly evident in the ASR rate for CTRL and BadCLIP, highlighting their dependence on end-to-end testing. Regarding ASSET, it demonstrates defensive utility against BadEncoder, CTRL, and DRUPE; however, when compared to our approach, its defense capability is approximately 40\% lower in terms of attack success probability. Furthermore, ASSET completely fails to defend against CLIP attacks, whereas we achieve a 100\% success rate in defending against CLIP Backdoor and a 70\% success rate in defending against BadCLIP. Though ASSET demonstrates the detection ability to detect BadCLIP to some extent in Table \ref{table: upstream}, but CLIP backdoor is effective as long as a few images pass the sifting. 

\begin{figure}[t]
  \centering
  \vspace{-6mm}
   \includegraphics[width=\linewidth]{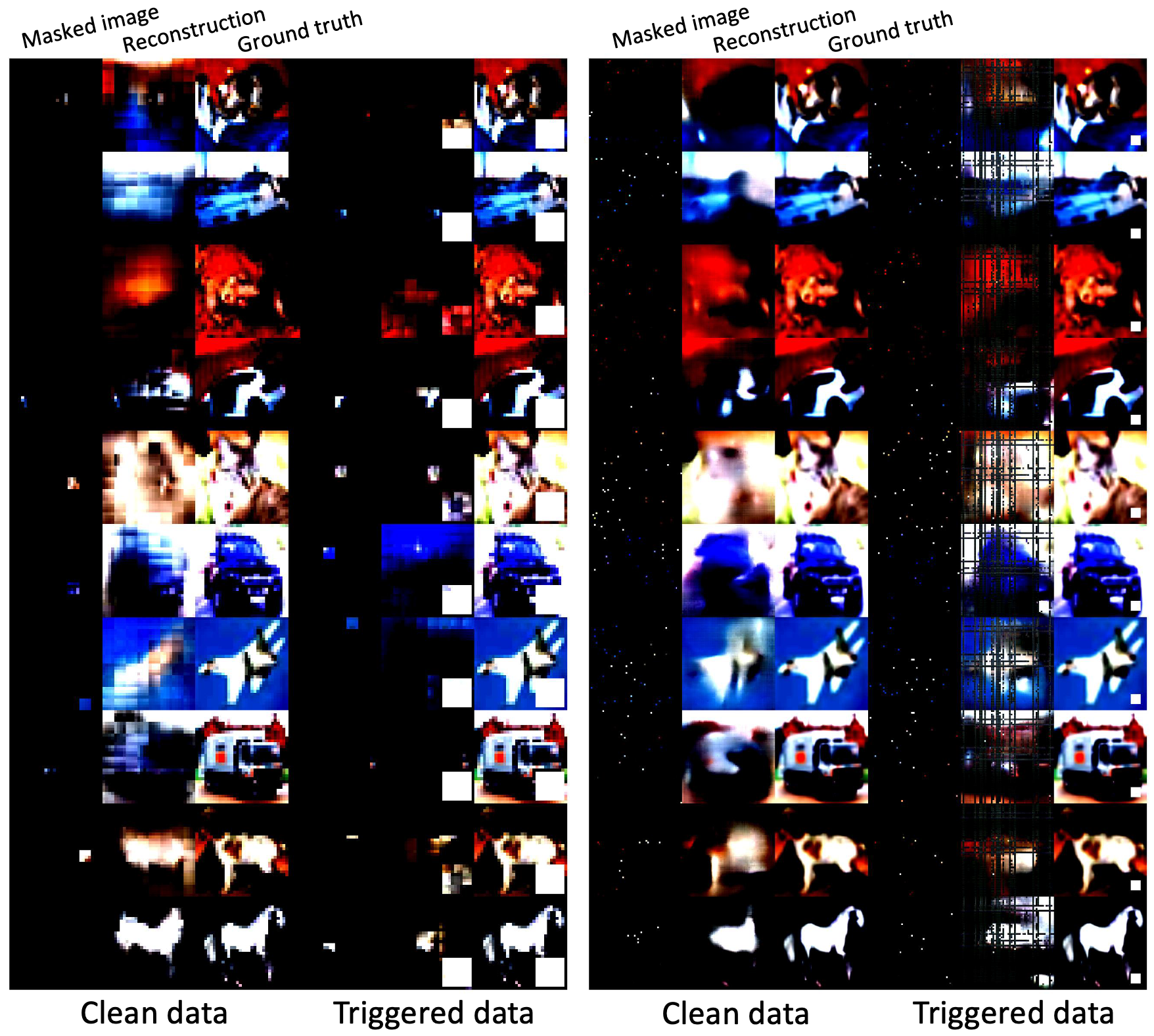}
   \vspace{-5mm}
   \caption{Reconstruction of \scheme. The left side and the right side are attacked by DRUPE and CLIP-Backdoor respectively. In six columns, the left three are for clean inputs and the right three are for poisoned inputs. In three columns, they are masked image, reconstructed image, and ground truth (input) image accordingly.}
   \vspace{-5mm}
   \label{fig: recon demo}
   
\end{figure}

We also present two representatives of DRUPE and CLIP-backdoor for the decoding results in Figure \ref{fig: recon demo}. In the images for DRUPE, it is noticeable that all reconstructed images have the trigger pattern, suggesting their embeddings are highly corrupted towards a backdoor distribution leading to high ASR presented in Table \ref{table: upstream}. Also as a result, our reconstruction effectively reverses the corrupted embeddings, leading to a significant error when compared to the input images. For CLIP images, the reversed images lose grid-shaped pixel information due to the corrupted embeddings. Although the remaining pixels of the reconstruction are similar, it still results in a sufficiently large loss that distinguishes backdoor samples in the test dataset.
More results for different dataset, unbalanced data distribution and reconstruction demonstrations for frequency-based/imperceptible attacks are presented in Appendix.














\vspace{-2mm}

\section{Conclusion}
\vspace{-2mm}

In this paper, we propose an SSL backdoor detection method \scheme that has no reliance on knowledge of the training dataset or the trigger type and is non-invasive to the victim encoder. Our empirical study demonstrates the \scheme's detection capability among different learning paradigms of CL and CLIP, as well as more stealthy backdoor attacks, hence, improving the scarcity and limitations in SSL backdoor countermeasures.

\newpage

{
    \small
    \section*{Acknowledgement}
  This work is in part supported by the Fundamental Research Funds for the Central Universities (Grant No. 2242024k30059).

}

\clearpage

\appendix

\section*{Appendix}

\section{Additional Visualizations}

\begin{figure*}[bp]
  \centering
  \includegraphics[width=1.03\linewidth]{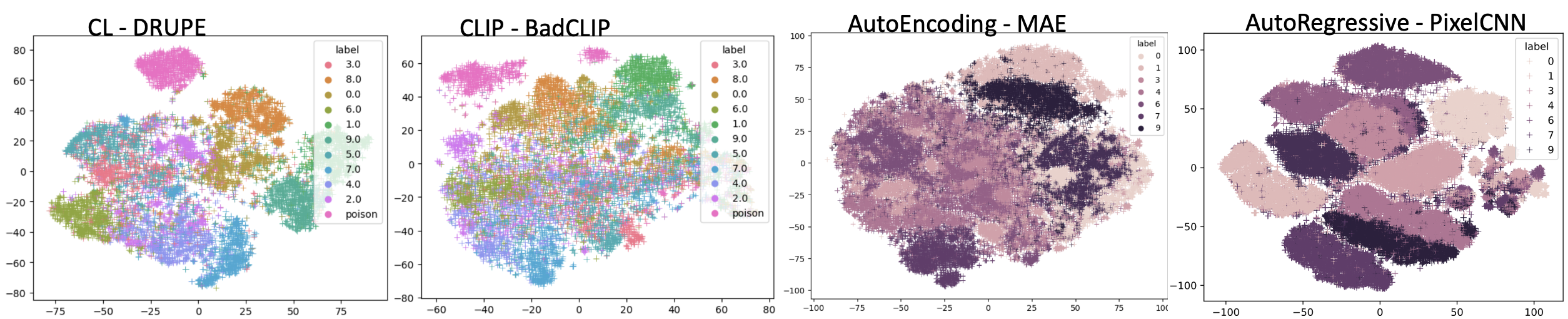}
  \vspace{-8mm}
   \caption{t-SNE visualization of embeddings in \textbf{different SSLs}.}
   \vspace{-6mm}
   \label{fig:tsne}
\end{figure*}

There is a common property in embedding distribution: \textit{images with similar features tend to cluster together in SSL}. We present t-SNE plots for better visualization of different SSL paradigms in Fig.\ref{fig:tsne}, including CL, CLIP, MAE, and auto-regressive model. With no exception, samples from the same classes are clustered in the embedding space. It serves a fundamental fact in various SSL models, which we have leveraged when designing DeDe. 

\section{Additional Experimental Results}
\begin{itemize}
    \item Table \ref{table: upstream2}: Upstream detection results for unbalanced data.
    \item Table \ref{table: downstream2}: Downstream defense performance for GTSRB.
    \item Table \ref{table: downstream3}: Downstream defense performance for  SVHN.
    \item Table \ref{table: recon2}: The reconstruction examples and error distributions, in which different parameter settings are presented. 
\end{itemize}
As a reminder, the balanced dataset presented in the paper is half poisoned (50\%) test dataset of size $10000$. Table \ref{table: upstream2} presents the result of a slightly poisoned (1\%) test dataset of the same size. ASSET's training dataset is kept with the same poisoning rate as the test dataset.
In the context of upstream detection, the observed trend aligns with what is presented in balanced data. Although DECREE demonstrates low AUC scores, it successfully identifies backdoor attacks in BadEncoder and CLIP-Backdoor. ASSET is capable of detecting BadEncoder and CTRL attacks but is ineffective against the stealthy DRUPE and CLIP-Backdoor attacks. Notably, ASSET shows strong performance in BadCLIP, suggesting its sensitivity to the selected poisoning rate. In contrast, \scheme maintains consistent performance across all attacks.
The downstream defense performance shows a decrease relative to the performance on CIFAR10, as reported in the main text. In comparison, ASSET demonstrates defense capabilities against CL attacks but is ineffective against CLIP attacks. Conversely, \scheme achieves an approximately $40\%$ improvement over all CL attacks and consistently defends against CLIP attacks. Although the defense performance in BadCLIP is not as strong as in other cases, it still reduces the attack success rate to 30\% in both scenarios.

In Table \ref{table: recon2}, we present reconstruction examples for all attacks using our method, \scheme. We use the same samples to present consistent visualization, so the images for CLIP and BadCLIP are up-sampled to $224\times 224$ for demonstration.  To present the robustness in comparison to ASSET, we present the error histograms of both methods, as they are both unsupervised techniques for detecting backdoor samples by computing losses. ASSET effectively distinguishes between clean and backdoor samples in the case of BadEncoder, successfully separating the two modes. However, its performance declines when it struggles to differentiate the backdoor samples in the first place, resulting in a mixing of the two modes. In contrast, while \scheme does not push the discerned samples further, it demonstrates sufficient strength to create two distinct modes, allowing for the use of a threshold to filter samples. It is also worth noting, that the results of DeDe are generally stable for different choices of patch size and masking ratio. It is reasonable to choose masking ratio in the range of $[0.75,0.95]$, which is supported by our testing results.

\textbf{DeDe training overhead}. In training the DeDe decoder, the given encoder(poisoned) is frozen for inference. Learning is on ViT-B/16 model as the decoder. We use a machine with Intel(R) Xeon(R) Gold 5118 CPU@2.30GHz and NVIDIA GeForce RTX 4090 GPU. Taking DRUPE as an example, max epoch is set to 200 and DeDe training dataset size is 50k. The total run time is 1hr'11m'30s, which is around 20 s/epoch. The training time is generally consistent in different attacks while experiments with $224\times 224$ image dataset take a bit longer.

\begin{table*}[hbtp]
\vspace{-5mm}
\caption{Upstream Detection Performance for unbalanced data.}
\vspace{-3mm}
\label{table: upstream2}
\centering
\scriptsize

\begin{tabular}{cccccccccccccccc}\toprule
& \multicolumn{3}{c}{BadEncoder} & \multicolumn{3}{c}{CTRL}& \multicolumn{3}{c}{DRUPE}& \multicolumn{3}{c}{CLIP Backdoor}& \multicolumn{3}{c}{BadCLIP}
\\\cmidrule(lr){2-4}\cmidrule(lr){5-7}\cmidrule(lr){8-10} \cmidrule(lr){11-13} \cmidrule(lr){14-16}
           & \textit{TPR} $(\uparrow)$  & \textit{FPR}$(\downarrow)$ & \textit{AUC}$(\uparrow)$    & \textit{TPR}  & \textit{FPR} & \textit{AUC} & \textit{TPR}  & \textit{FPR} & \textit{AUC}& \textit{TPR}  & \textit{FPR} & \textit{AUC} & \textit{TPR}  & \textit{FPR} & \textit{AUC} \\\midrule
DECREE    & 66.0$^*$  & 42.1$^*$ & 0.336$^*$   & - &  -  & -  & 68.0 & 49.8 & 0.366 & 70.0$^*$ & 49.7$^*$ & 0.363$^*$ & - &  -  & - \\
ASSET &  100.0 & 25.0 & 0.901 & 31.1 & 91.0  & 0.799 & 94.8 & 27.6 & 0.858 &  54.4 & 47.8 & 0.555 & 100.0 & 25.0 & \textbf{0.943} \\
DEDE & 92.0 & 8.3 & 0.978 & 90.0 & 19.3 & \textbf{0.898} & 98.5 & 3.1 & \textbf{0.998} & 100.0 & 0.0 & \textbf{1.0}  & 84.0 & 19.4 & 0.903 \\
DEDE OOD & 96.5 & 8.2  & \textbf{0.983}  & 81.0 & 19.3 & 0.853 &  95.5 & 8.3 & 0.979 & 100.0 & 0.0 & \textbf{1.0 } & 88.0 & 19.3 & \textbf{0.936} \\\bottomrule

\end{tabular}
\end{table*}

\begin{table*}[hbtp]
\caption{Downstream Performance for GTSRB.}
\vspace{-3mm}
\label{table: downstream2}
\centering
\footnotesize
\begin{tabular}{ccccccccccccc}\toprule
& \multicolumn{2}{c}{No Attack} &  \multicolumn{2}{c}{BadEncoder} & \multicolumn{2}{c}{CTRL}& \multicolumn{2}{c}{DRUPE}& \multicolumn{2}{c}{CLIP Backdoor}& \multicolumn{2}{c}{BadCLIP}
\\  \cmidrule(lr){2-3} \cmidrule(lr){4-5} \cmidrule(lr){6-7} \cmidrule(lr){8-9} \cmidrule(lr){10-11} \cmidrule(lr){12-13} & \textit{CA} $(\uparrow)$ & \textit{ASR} $(\downarrow)$ & \textit{CA} & \textit{ASR}& \textit{CA} & \textit{ASR}& \textit{CA} & \textit{ASR}& \textit{CA} & \textit{ASR}& \textit{CA} & \textit{ASR}\\\midrule
No Def. Poison &  85.14 & 19.16 & 82.03 & 99.36 & 73.9&  91.57 & 85.6&  99.46 & 73.76 & 97.24 & 76.99 & 98.59 \\ No Def. Clean & 85.25  & 10.85 &  82.26  & 97.27 &  67.48 &  64.26 &  81.60 &  97.03 &  74.51  & 98.05  & 79.74  & 11.69 \\ ASSET & -  & - &  83.24  & 54.67 &  67.45 &  50.51  & 80.85  & 54.07  & 73.92  & 96.42  & 76.19 & 98.38 \\  DEDE & -  & -  & 82.86 &  2.99  & 68.61 &  4.43  & 80.46  & 0.91  & 74.41  & 2.21 & 76.20 & 30.60
    \\\bottomrule
\end{tabular}
\end{table*}

\begin{table*}[hbtp]
\caption{Downstream Performance for SVHN.}
\vspace{-3mm}
\label{table: downstream3}
\centering
\footnotesize
\begin{tabular}{ccccccccccccc}\toprule
& \multicolumn{2}{c}{No Attack} &  \multicolumn{2}{c}{BadEncoder} & \multicolumn{2}{c}{CTRL}& \multicolumn{2}{c}{DRUPE}& \multicolumn{2}{c}{CLIP Backdoor}& \multicolumn{2}{c}{BadCLIP}
\\  \cmidrule(lr){2-3} \cmidrule(lr){4-5} \cmidrule(lr){6-7} \cmidrule(lr){8-9} \cmidrule(lr){10-11} \cmidrule(lr){12-13} & \textit{CA} $(\uparrow)$ & \textit{ASR} $(\downarrow)$ & \textit{CA} & \textit{ASR}& \textit{CA} & \textit{ASR}& \textit{CA} & \textit{ASR}& \textit{CA} & \textit{ASR}& \textit{CA} & \textit{ASR}\\\midrule
  No Def. Poison &  84.22 & 17.43 & 79.97 & 99.77 & 73.9&  88.22 & 85.6&  97.41 & 73.03 & 98.44 & 75.31 & 98.01 \\ No Def. Clean & 85.38  & 10.34 &  78.25  & 99.70 &  67.84 &  63.29 &  77.56 &  96.86 &  73.89  & 97.23  & 75.45  & 11.57 \\ ASSET & -  & - &  81.09  & 56.86 &  66.52 &  48.95  & 79.26  & 53.45  & 70.50  & 99.39  & 73.88 & 95.80 \\  DEDE & -  & -  & 79.28 &  1.53  & 64.70 &  4.78  & 80.03  & 1.85  & 73.93  & 1.82 & 75.12 & 30.51
    \\\bottomrule
\end{tabular}
\end{table*}

\begin{table*}[hbp]
\caption{Reconstruction results of \scheme. The three rows are reconstruction plots, \scheme 's detection error histogram, and ASSET's detection error histogram. For the reconstruction plots, six columns are $\{$masked image, reconstruction result, ground truth$\}$ for clean images and $\{$masked image, reconstruction result, ground truth$\}$ for backdoor images respectively.}
\label{table: recon2}
\begin{tabular}{cccccc}\toprule
& BadEncoder & CTRL & DRUPE & CLIP  & BadCLIP \\
\cmidrule(lr){2-2} \cmidrule(lr){3-3} \cmidrule(lr){4-4} \cmidrule(lr){5-5} \cmidrule(lr){6-6} 
\begin{turn}{90} 
Reconstruction Examples
\end{turn}  
& \includegraphics[width=0.18\textwidth]{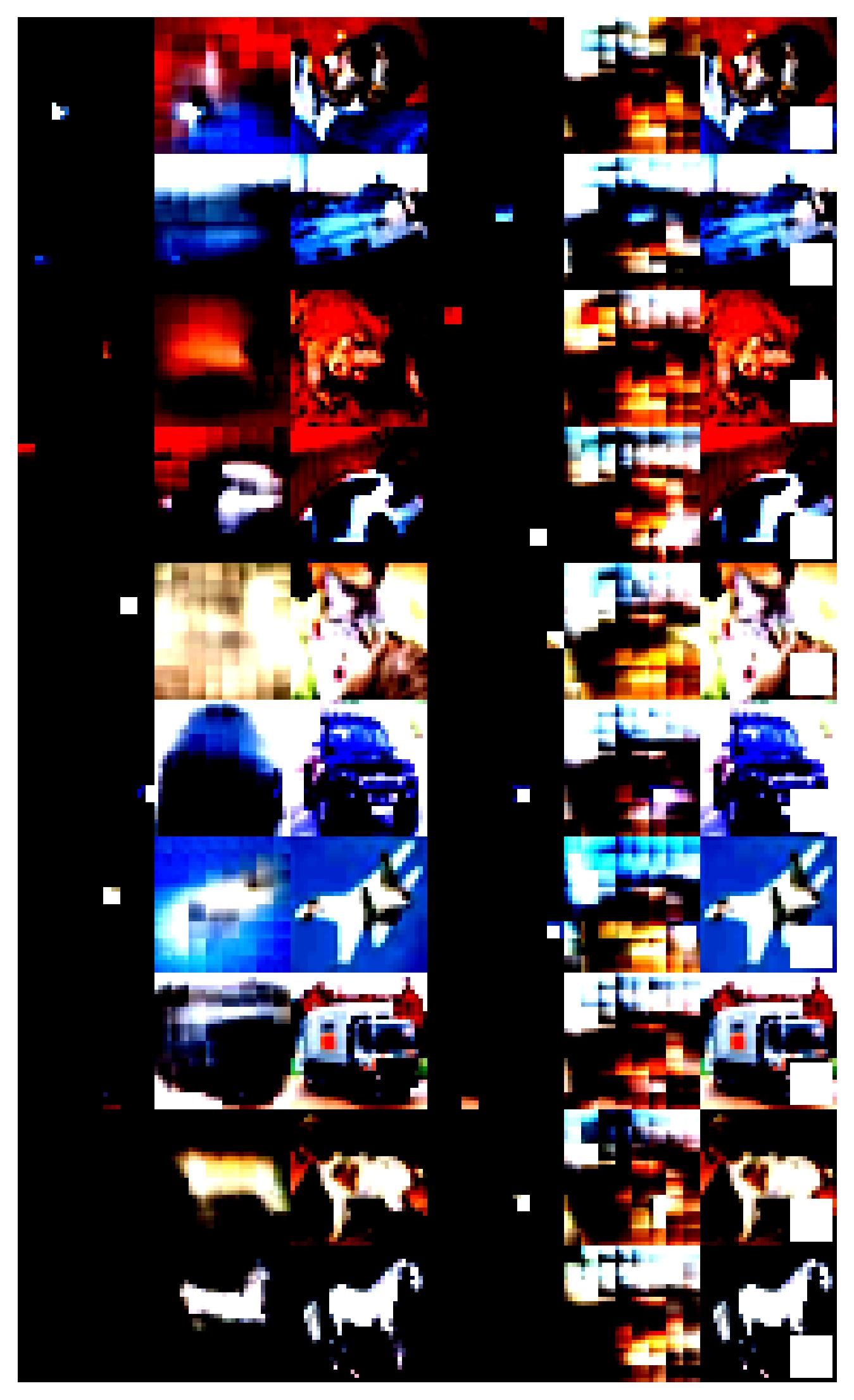}     & \includegraphics[width=0.18\textwidth]{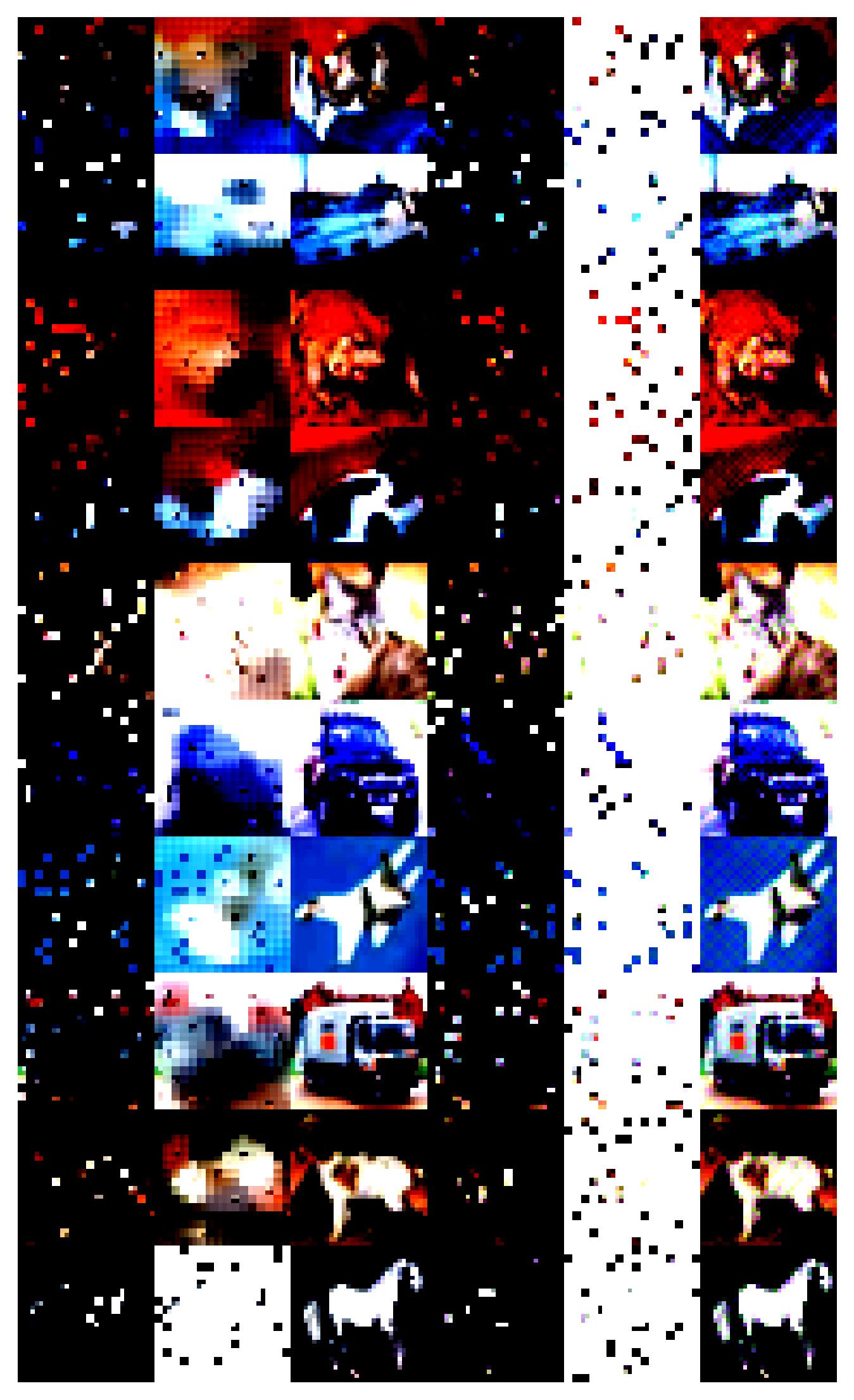}     &   \includegraphics[width=0.18\textwidth]{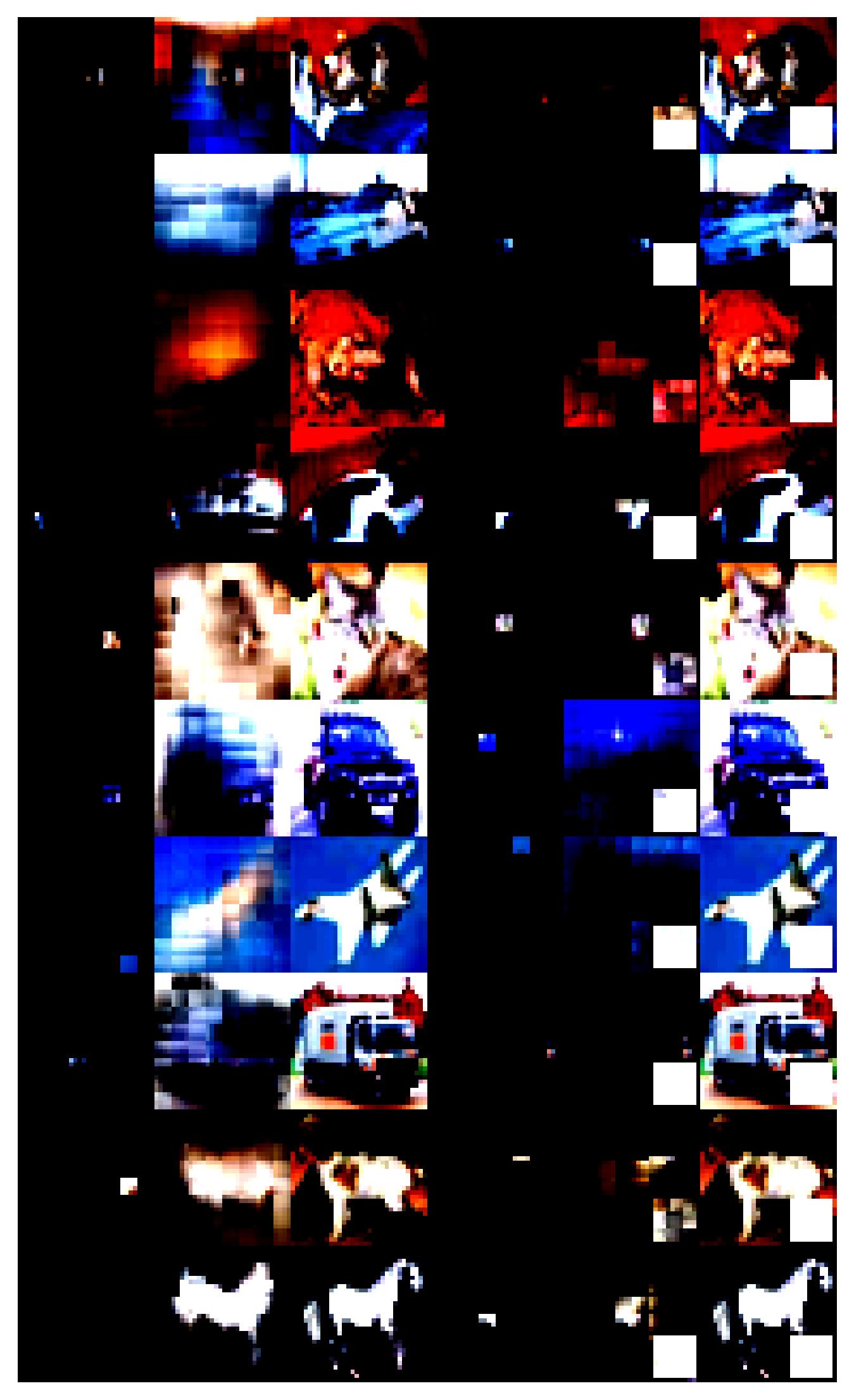}     &   \includegraphics[width=0.18\textwidth]{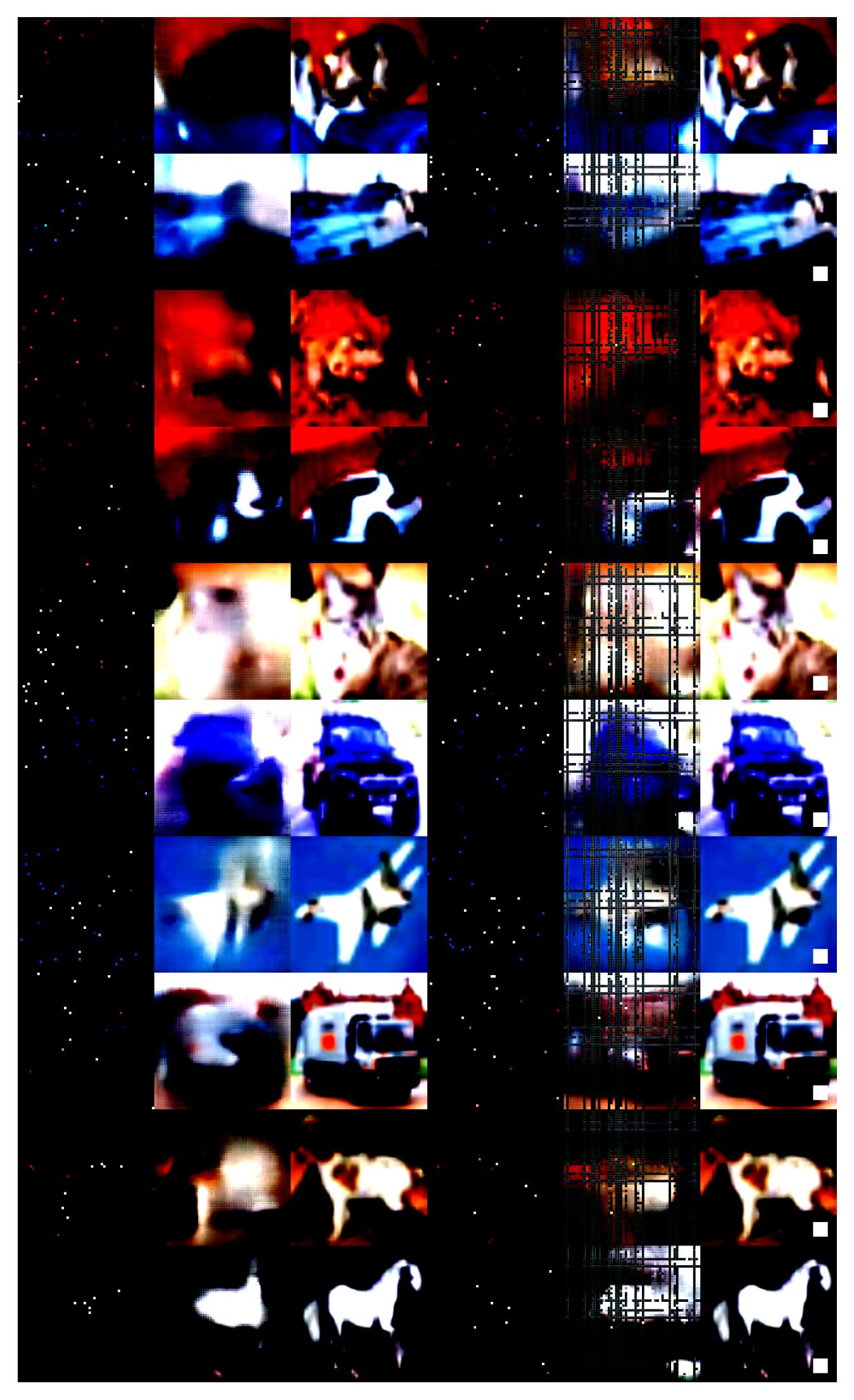}    &    \includegraphics[width=0.18\textwidth]{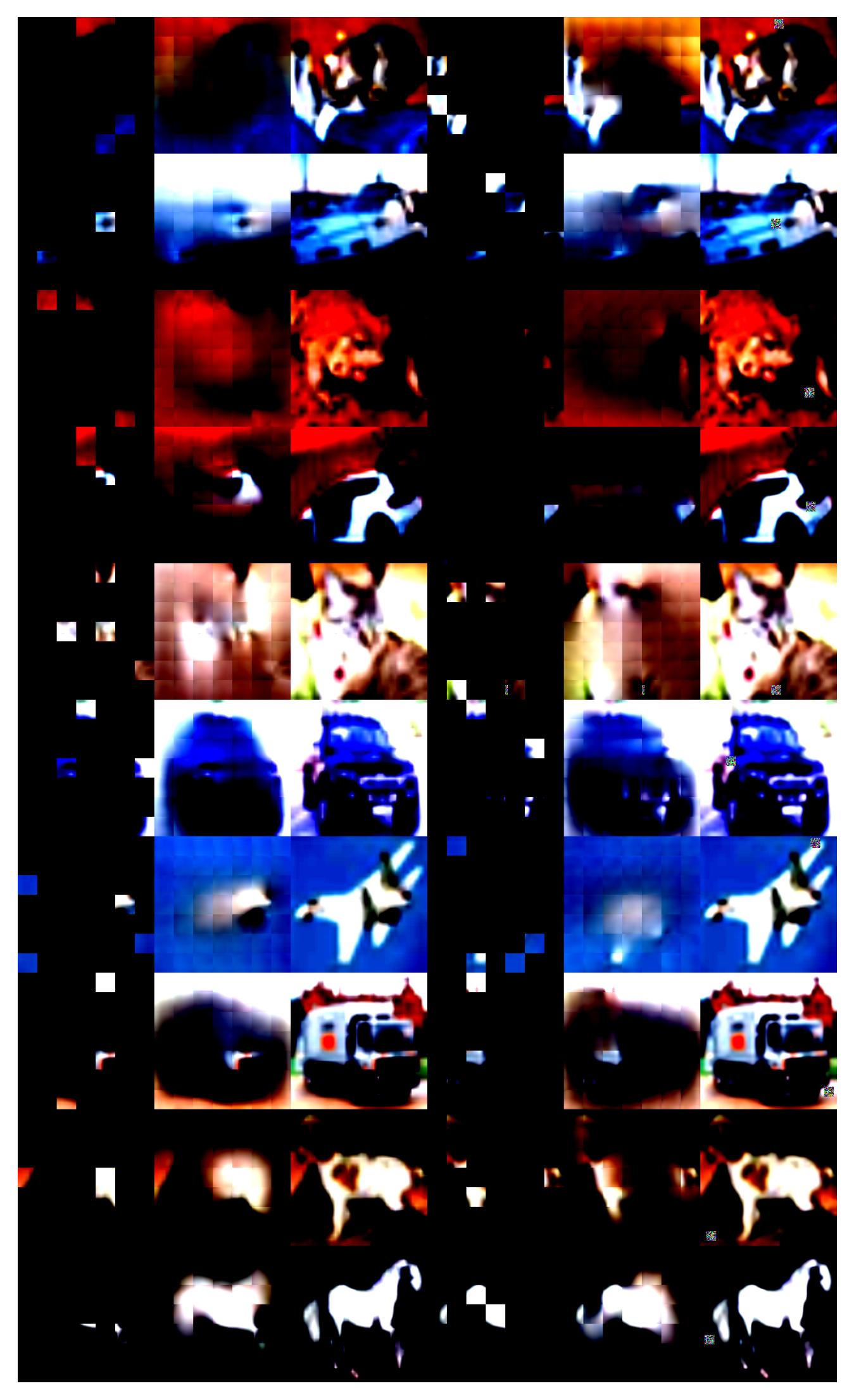}     \\
\begin{turn}{90} 
\scheme Dist.  
\end{turn}  
&
\includegraphics[width=0.18\textwidth]{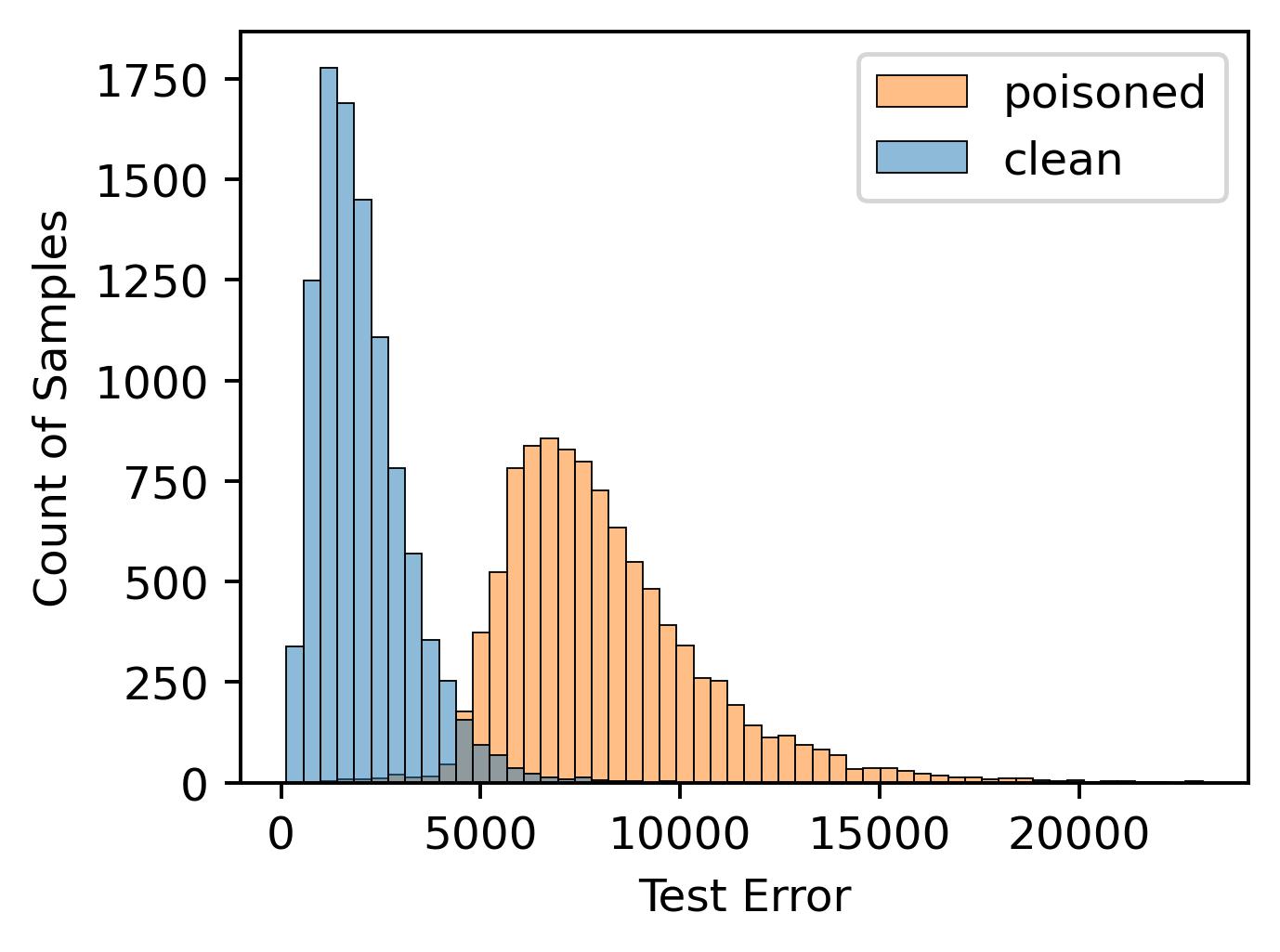}        &  \includegraphics[width=0.18\textwidth]{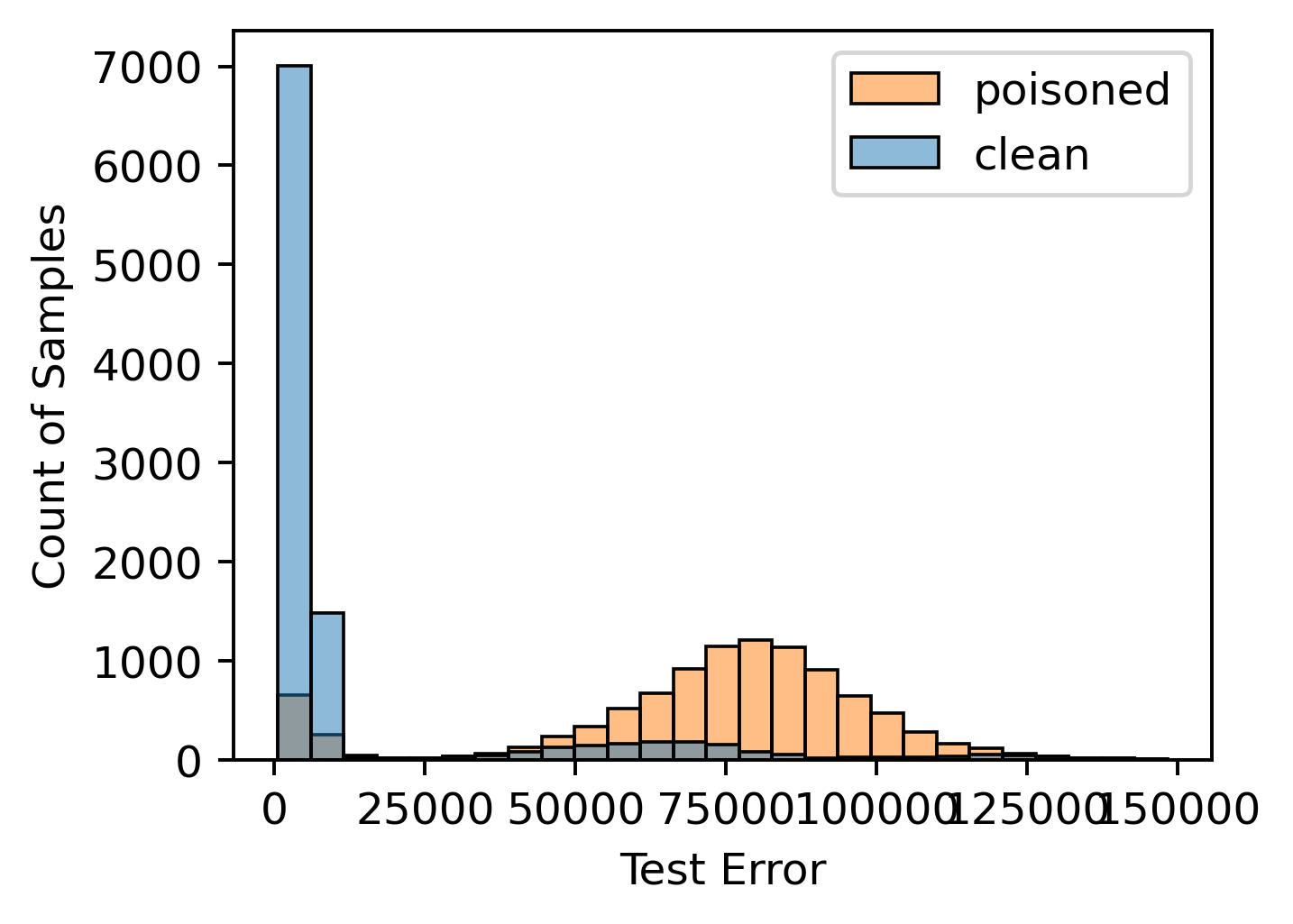}     &    \includegraphics[width=0.18\textwidth]{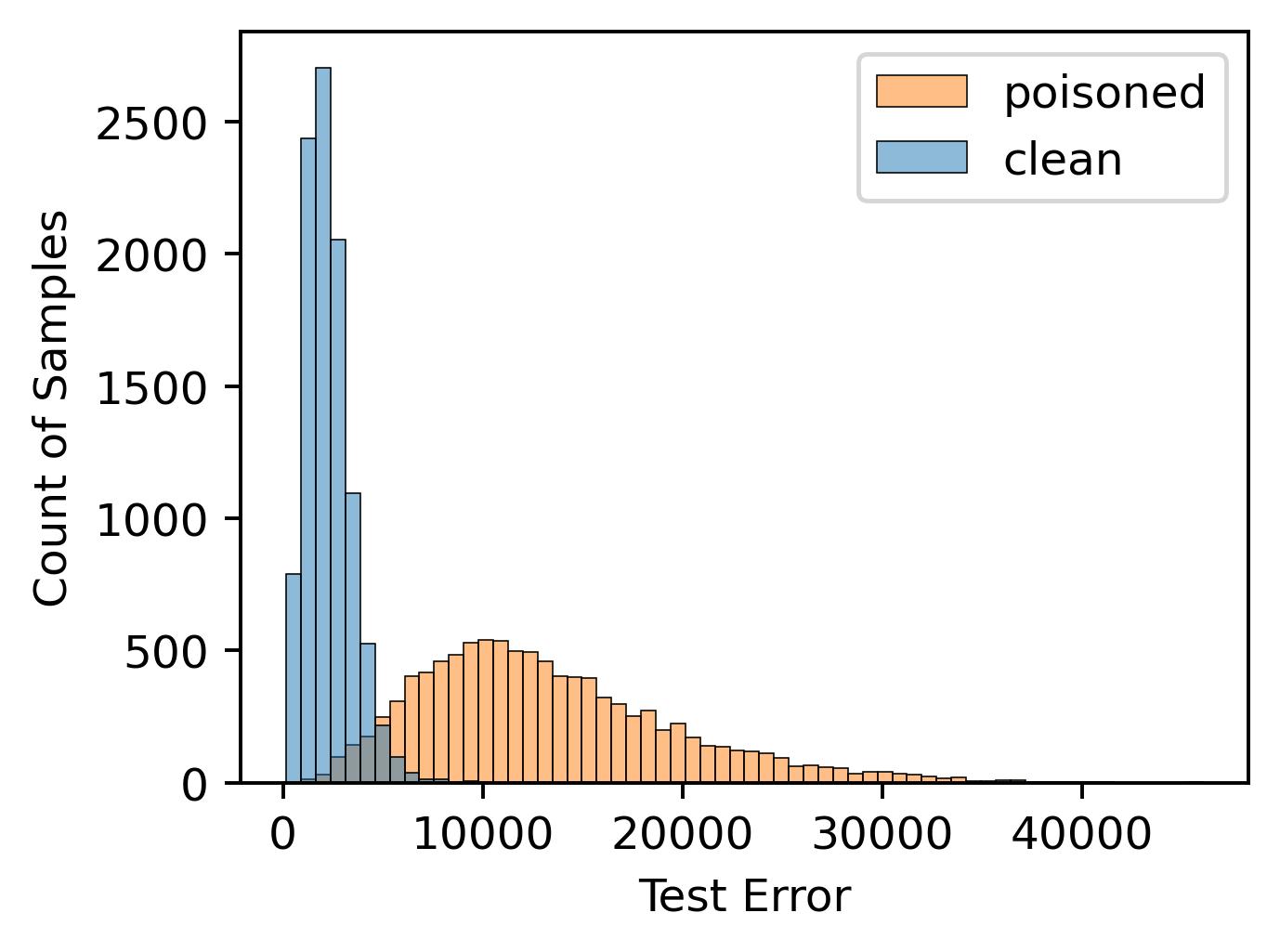}    & \includegraphics[width=0.18\textwidth]{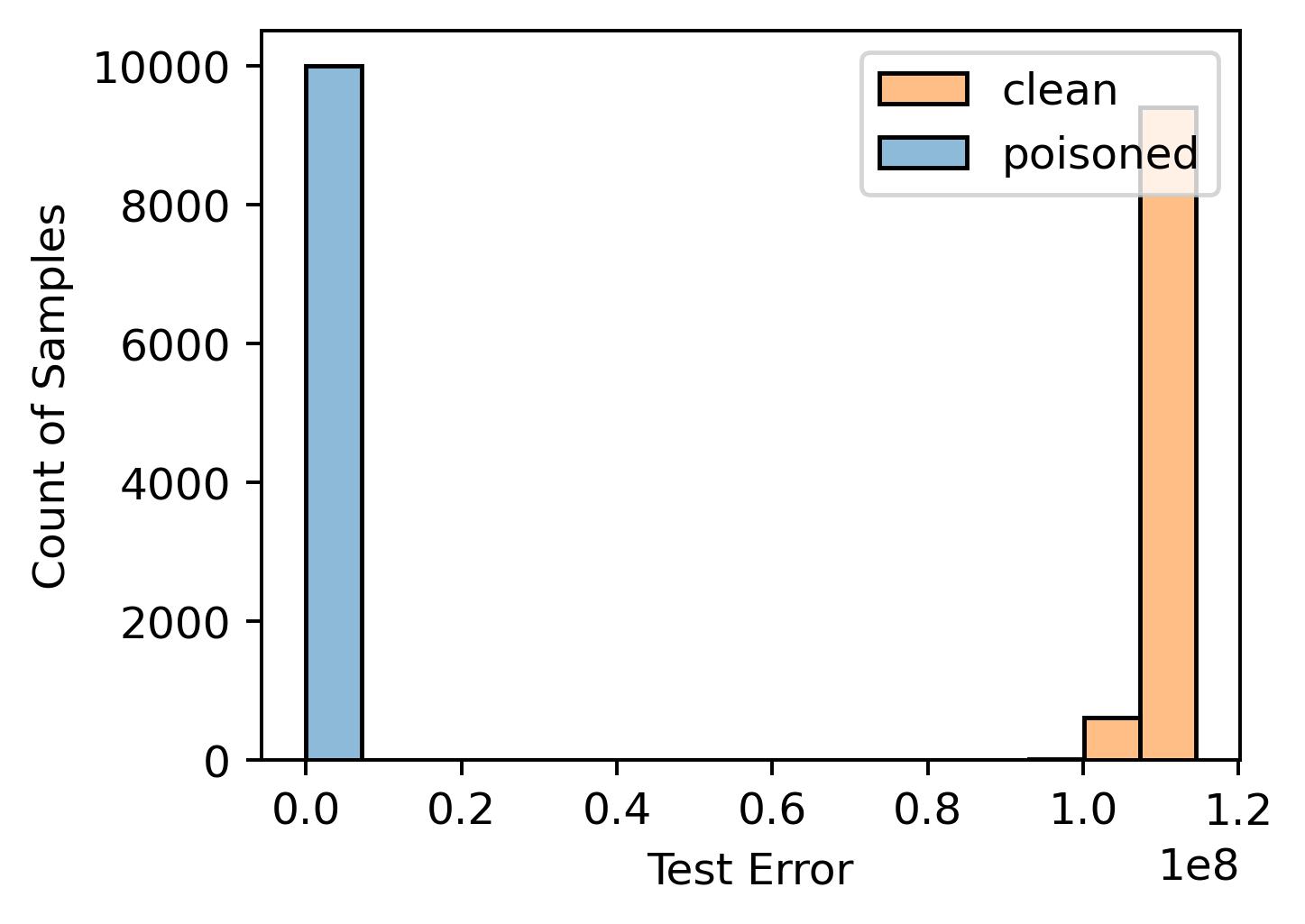}      &   \includegraphics[width=0.18\textwidth]{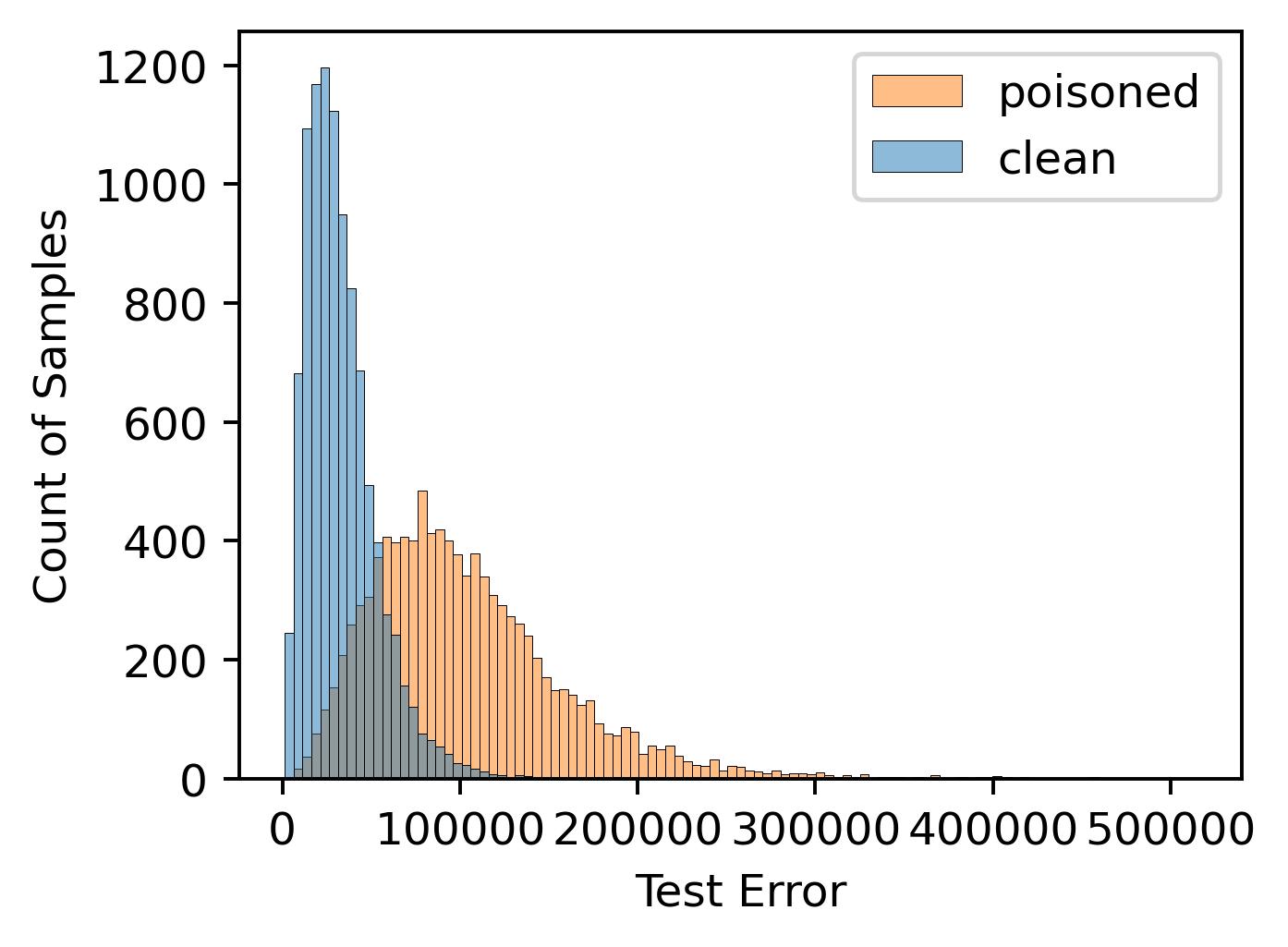}   
\\
\begin{turn}{90} 
Asset Dist.
\end{turn}  
&
\includegraphics[width=0.18\textwidth]{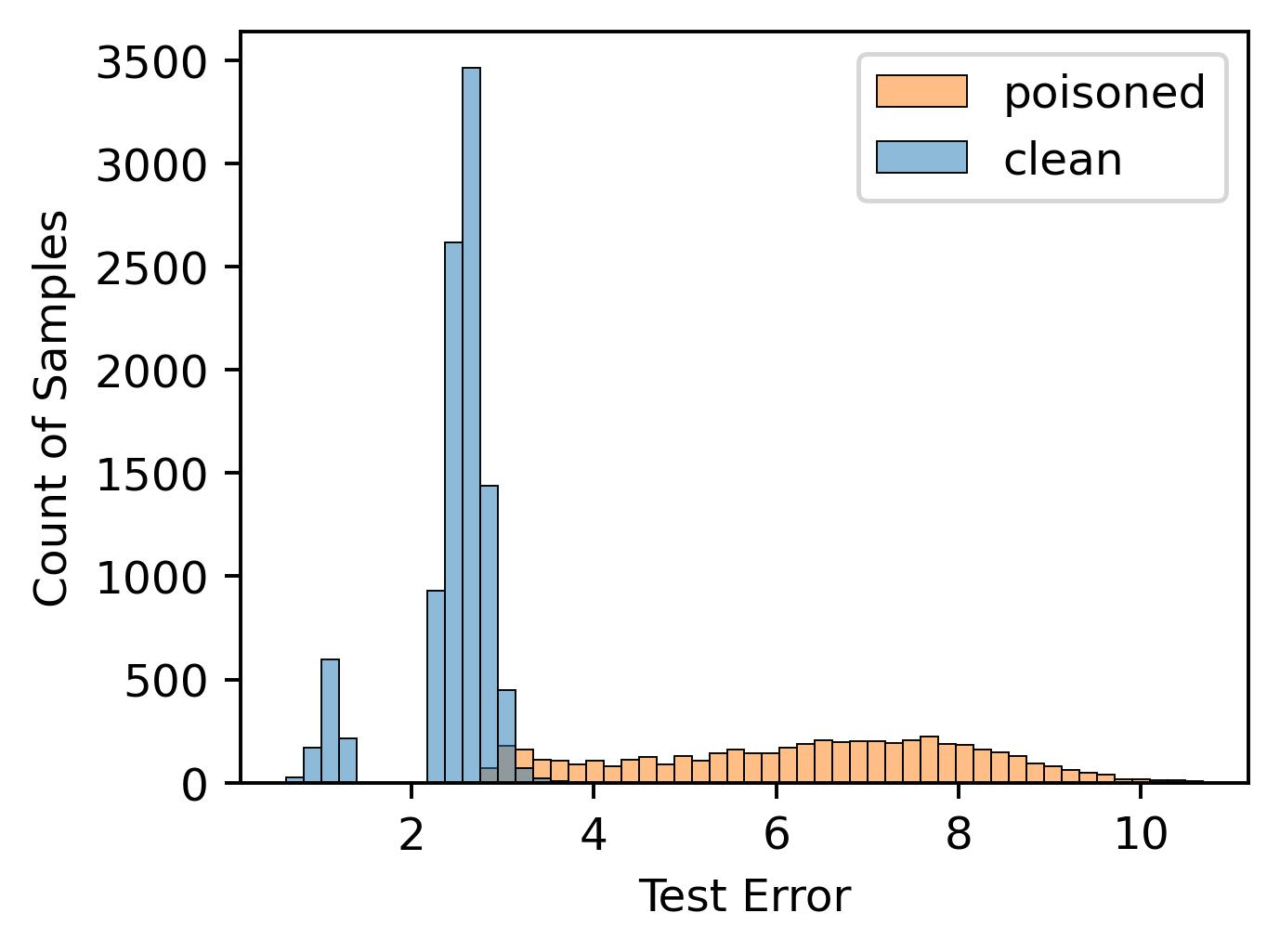}        &   \includegraphics[width=0.18\textwidth]{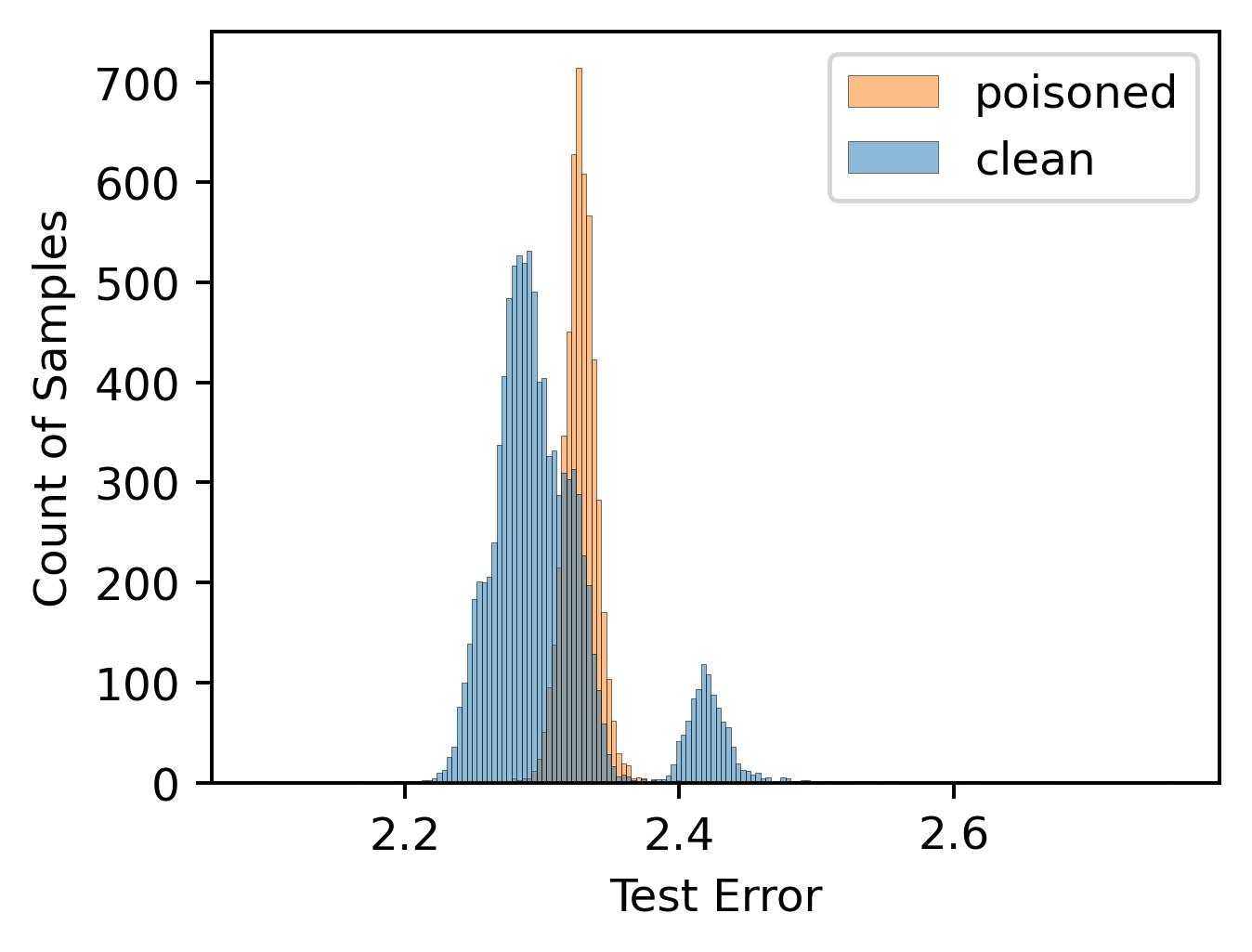}   &     \includegraphics[width=0.18\textwidth]{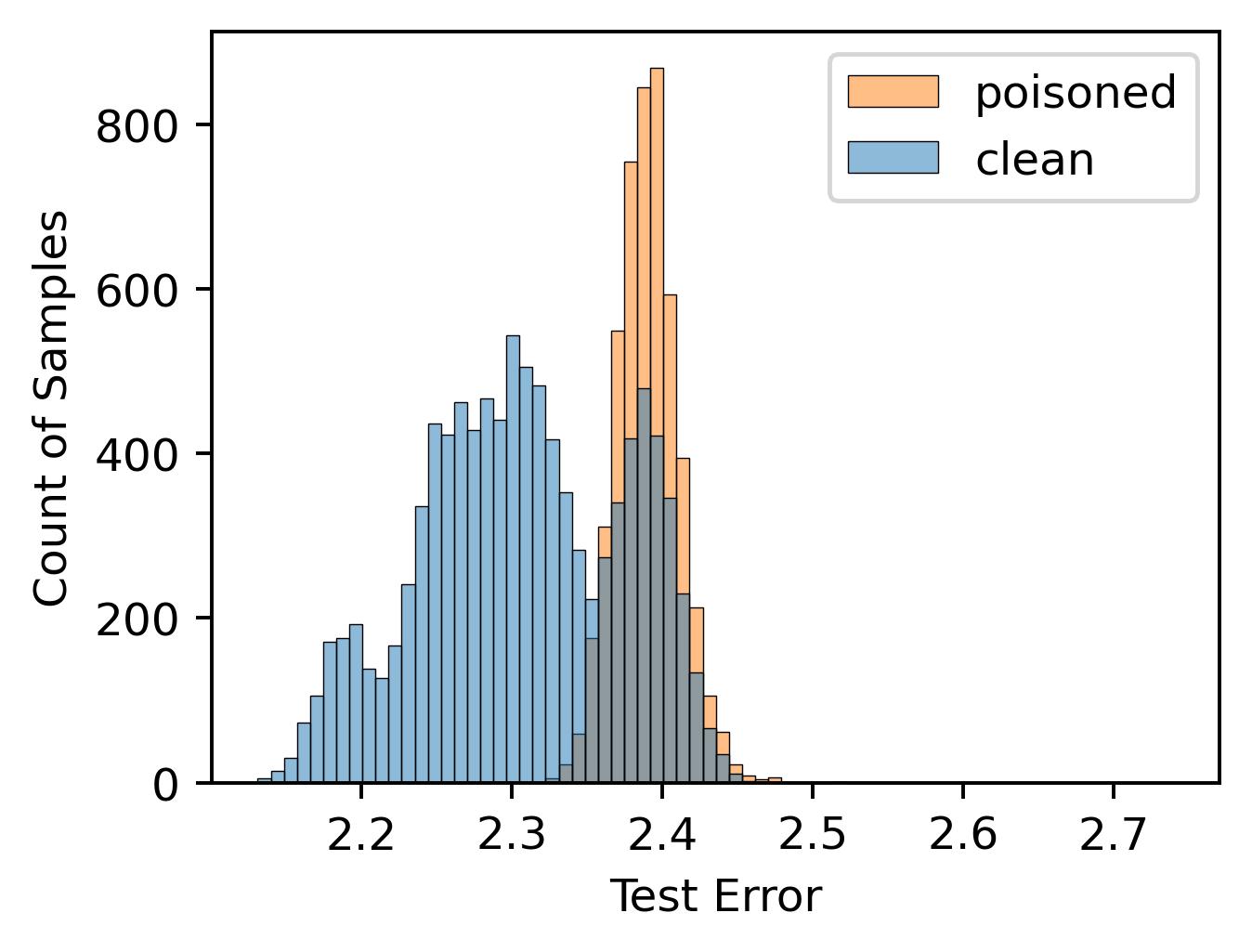}  &   \includegraphics[width=0.18\textwidth]{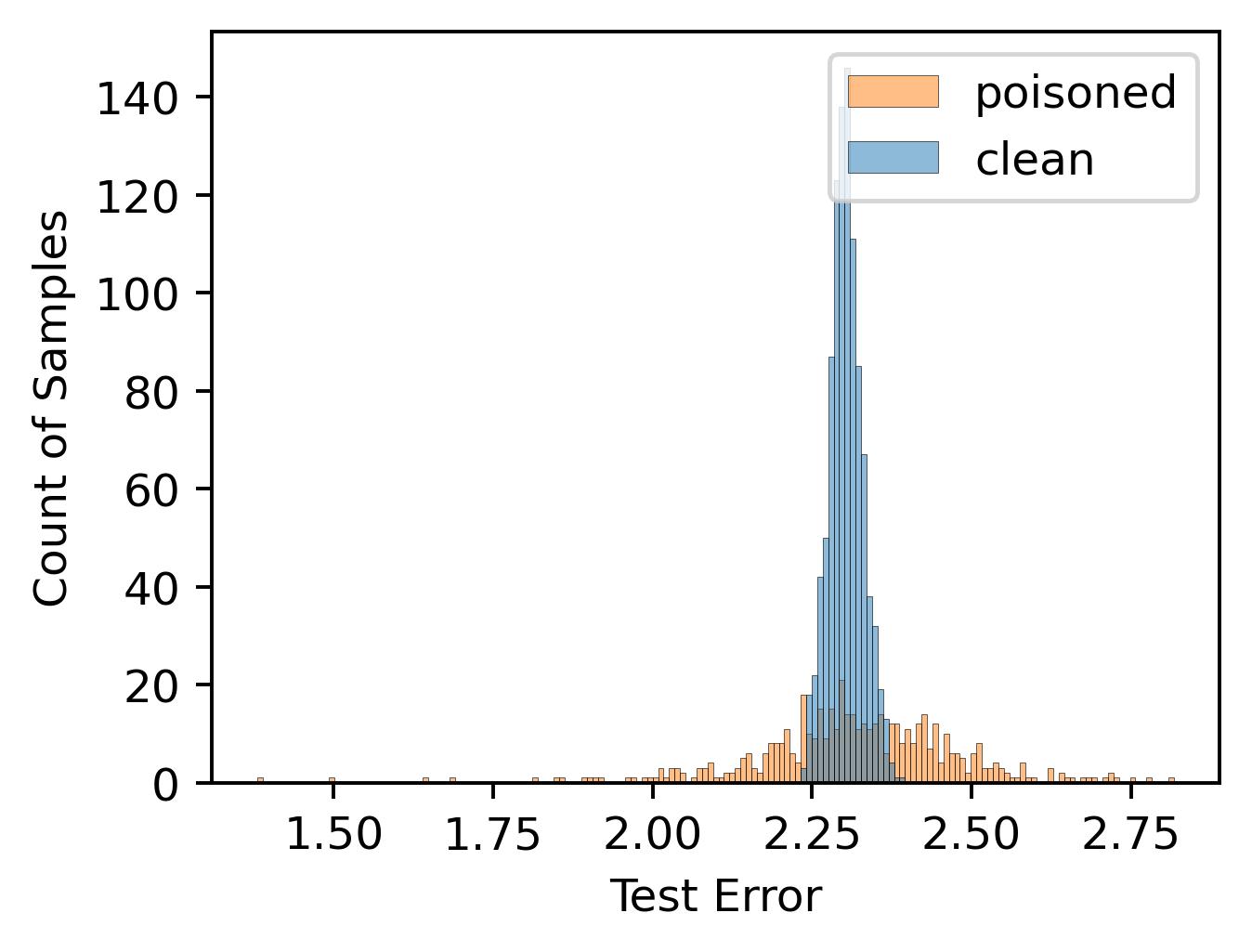}   &    \includegraphics[width=0.18\textwidth]{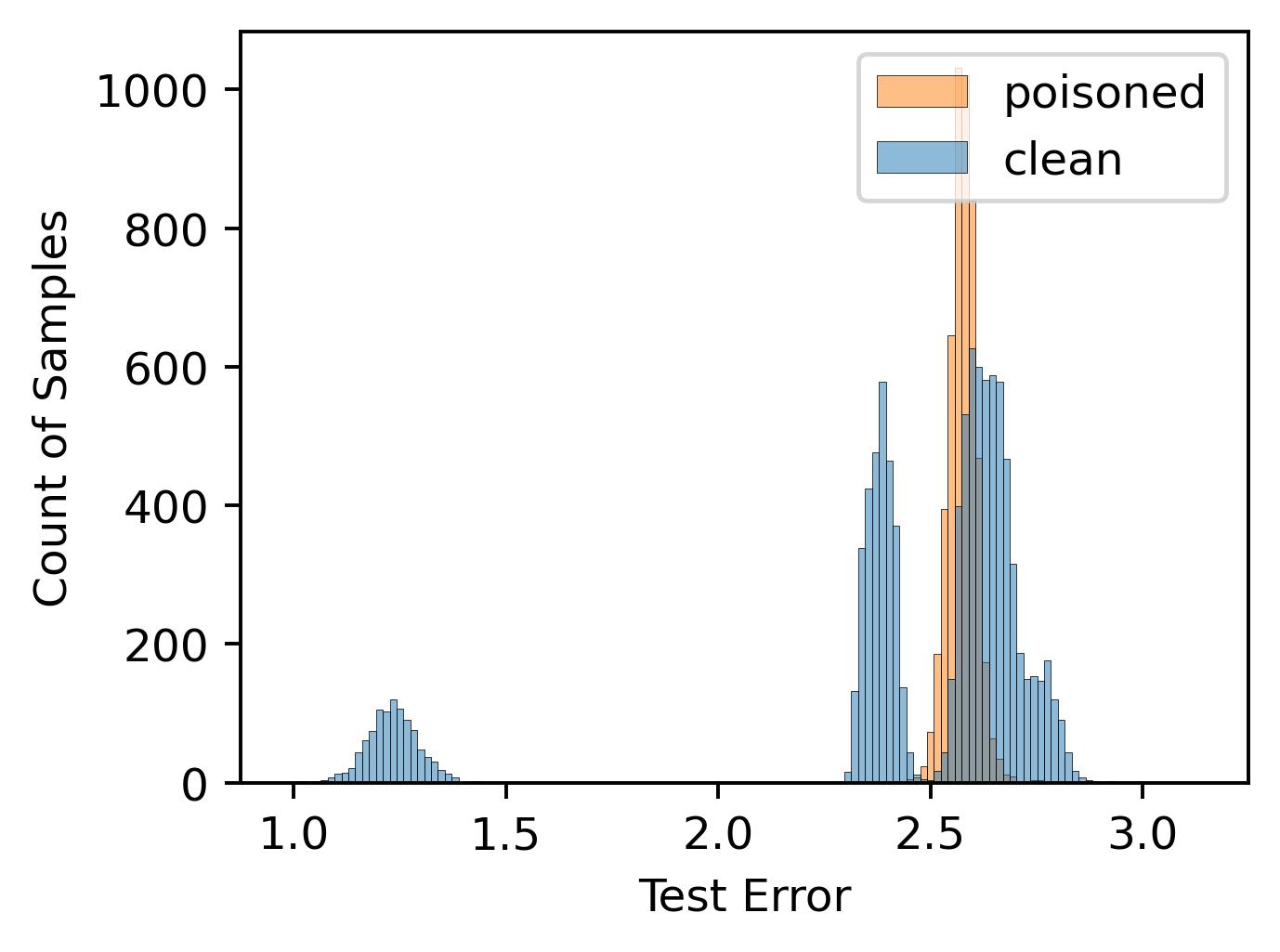}  

\\\bottomrule
\end{tabular}

\end{table*}


\end{document}